\documentclass[letterpaper]{article} 
\usepackage{aaai2026}  
\usepackage{times}  
\usepackage{helvet}  
\usepackage{courier}  
\usepackage[hyphens]{url}  
\usepackage{graphicx} 
\usepackage{natbib}  
\usepackage{caption} 
\usepackage{algorithm}
\usepackage{algorithmic}

\usepackage{newfloat}
\usepackage{listings}
\DeclareCaptionStyle{ruled}{labelfont=normalfont,labelsep=colon,strut=off} 
\floatstyle{ruled}
\newfloat{listing}{tb}{lst}{}
\floatname{listing}{Listing}
\usepackage{graphicx}
\usepackage{tabularx}
\usepackage{booktabs}
\usepackage{multirow}
\usepackage{adjustbox}
\usepackage{array} 
\usepackage{amssymb} 
\usepackage{makecell}
\usepackage{pifont}
\usepackage{url}
\usepackage{amsmath}
\usepackage{subfigure}
\usepackage{tcolorbox}
\tcbuselibrary{breakable}
\usepackage{caption}
\newcommand{\homa}[1]{\textcolor{blue}{H: #1}}

\title{Expert-Guided Prompting and Retrieval-Augmented Generation for \\Emergency Medical Service Question Answering}
\author{
    Xueren Ge\textsuperscript{\rm 1},
    Sahil Murtaza\textsuperscript{\rm 1}, 
    Anthony Cortez\textsuperscript{\rm 2}, 
    Homa Alemzadeh\textsuperscript{\rm 1}
}
\affiliations{
    \textsuperscript{\rm 1}School of Engineering and Applied Sciences, University of Virginia\\
    \textsuperscript{\rm 2}School of Medicine, University of Virginia\\

    zar8jw@virginia.edu, vpn9ej@virginia.edu, aec3gp@virginia.edu, ha4d@virginia.edu
}

\usepackage{bibentry}

\begin{document}

\maketitle

\begin{abstract}
Large language models (LLMs) have shown promise in medical question answering, yet they often overlook the domain-specific expertise that professionals depend on-such as the clinical subject areas 
(e.g., trauma, airway) and the certification level (e.g., EMT, Paramedic). Existing approaches typically apply general-purpose prompting or retrieval strategies without leveraging this structured context, limiting performance in high-stakes settings. We address this gap with EMSQA, an 24.3K-question multiple-choice dataset
spanning 10 clinical subject areas and 4 certification levels, accompanied by curated, subject area-aligned knowledge bases (40K documents and 2M tokens).
Building on EMSQA, we introduce (i) Expert-CoT, a prompting strategy that conditions chain-of-thought (CoT) reasoning on specific clinical subject area and certification level, and (ii) ExpertRAG, a retrieval-augmented generation pipeline that grounds responses in subject area-aligned documents and real-world patient data. Experiments on 4 LLMs show that Expert-CoT improves up to 2.05\% over vanilla CoT prompting. Additionally, combining Expert-CoT with ExpertRAG yields up to a 4.59\% accuracy gain over standard RAG baselines. Notably, the 32B expertise-augmented LLMs pass all the computer-adaptive EMS certification simulation exams.

\end{abstract}
\begin{links}
\link{Code \& Data}{https://uva-dsa.github.io/EMSQA}
\end{links}

\section{Introduction}

The rapid advancement of large language models (LLMs) has brought new possibilities to high-stakes domains such as emergency medical services (EMS)~\cite{weerasinghe2024real}, where accurate and reliable decision-making is critical. There is growing interest in leveraging LLMs for medical education~\cite{abd2023large}, decision support~\cite{ge2024dkec, luo2025alpha}, and certification preparation~\cite{kung2023performance}, particularly in the context of open-domain multiple-choice question answering (MCQA). However, while LLMs have shown promising performance on general medical QA benchmarks~\cite{cai2024medbench}, important gaps remain between their current capabilities and the reasoning processes used by trained medical professionals.

Recent approaches in medical MCQA, such as chain-of-thought prompting (CoT)~\cite{wei2022chain} and retrieval-augmented generation (RAG)~\cite{lewis2020retrieval}, have improved LLM performance by enhancing reasoning capabilities and incorporating external domain knowledge during inference. But these approaches often treat both reasoning and retrieval as undifferentiated processes: the model sees a question and retrieves documents or reasons directly, without considering what kind of knowledge is relevant or how a human expert would approach the task. In contrast, real-world medical professionals typically begin by identifying the subject area of the question—e.g., whether it pertains to trauma, airway management, or pharmacology—and then reason from that domain-specific perspective, using knowledge and protocols appropriate to their level of certification.

Existing benchmarks such as MedQA~\cite{jin2021disease}, MMLU-Med~\cite{hendrycks2020measuring} and MedMCQA~\cite{pal2022medmcqa} lack both this structured representation of question expertise (e.g., subject area, certification level) and the associated domain-specific knowledge. This makes it difficult to align retrieval and reasoning processes with human-like problem-solving strategies. In particular, publicly available EMS question answering datasets and knowledge sources with expertise annotation are scarce. Further, current state-of-the-art (SOTA) methods do not account for how incorporating structured expertise can improve the overall effectiveness of reasoning and answer generation in retrieval-augmented systems.


\begin{figure*}[t!]
  \centering
  \includegraphics[width=\linewidth]{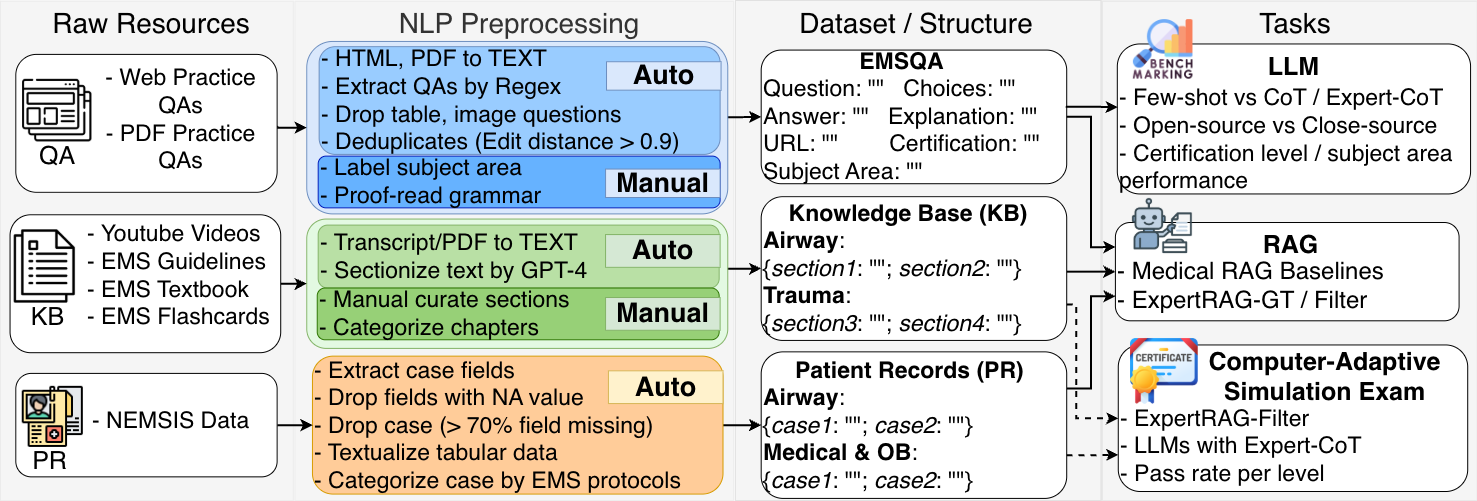}
  \caption{Overall Approach. (1) EMSQA, KB, PR Construction; (2) Tasks: Benchmark LLMs, RAG and Certification Exams.
  }
  \label{fig:Overall approach}
\end{figure*}

To address these gaps, we propose a new dataset and a \textit{domain-expertise-guided} LLM framework that models medical MCQA in a more structured, cognitively informed manner. As shown in Figure~\ref{fig:Overall approach}, our contributions are three-fold:
\begin{itemize}
    \item We introduce \textbf{EMSQA}, the first EMS MCQA dataset of 24.3K questions, curated based on public and private sources, covering 10 subject areas and 4 certification levels, and accompanied by a structured, subject area-aligned EMS knowledge base (KB) with 40K documents and 4M real-world patient care reports. Partial data (from public sources) and the whole EMS KB is shared as a resource with the EMS and research communities.
    \item We propose a \textit{expertise-guided} LLM framework that infers the domain expertise attributes and injects them into LLMs using two approaches: (1) an \textbf{expertise-guided prompting strategy (Expert-CoT)} that encourages step-by-step reasoning from a domain-specific perspective. (2) an \textbf{expertise-guided RAG method (ExpertRAG)} that selectively retrieves expertise-aligned knowledge from curated EMS KBs and patient records.
    \item We benchmark multiple LLMs on EMSQA, evaluating performance across certification levels and subject areas, and compare our framework against SOTA RAG methods. Experimental results show that combining \textbf{Expert-CoT} and \textbf{ExpertRAG} yields up to a 4.59\% improvement in accuracy. Notably, the 32B expertise‑augmented models pass all the EMS certification simulation exams.
\end{itemize}

\section{Related Works}
\subsection{Medical Question Answering Datasets}
Medical QA datasets typically fall into two paradigms: retrieval-based tasks~\cite{pampari-etal-2018-emrqa, krithara2023bioasq, ben2019question}, which require explicit evidence grounding by locating answers within documents, and open-domain multiple-choice tasks such as MedQA~\cite{jin2021disease}, MMLU-Med~\cite{hendryckstest2021}, and MedMCQA~\cite{pal2022medmcqa} that test implicit medical reasoning by choosing the best answer based on world knowledge. However, existing MCQA datasets focus on a single certification level and provide subject area labels without corresponding knowledge bases. EMSQA is the first open-domain medical MCQA dataset that spans multiple certification levels while also furnishing both subject area annotations and a structured EMS knowledge base. Table~\ref{tab:qa_comparison} shows a comparison between EMSQA and SOTA open-domain MCQA datasets.

\begin{table}[h]
  \centering
  \small
  \setlength{\tabcolsep}{1mm}
  \begin{tabular}{@{} l l l l @{}}
    \toprule
                 & \textbf{MedQA}   & \textbf{MedMCQA}   & \textbf{EMSQA}   \\
    \midrule
    Domain       & General Med.     & General Med.       & EMS              \\
    Data Size    & 12.7K           & 193K              & 24.3K     \\
    Exam         & USMLE            & AIIMS\&NEET PG   & NREMT            \\
    \#Certification & 1 & 1 & 4\\
    \#Subject Area     & \ding{55}        & 21                 & 10               \\
    KB           & Raw              & \ding{55}          & Categorized      \\
    \bottomrule
  \end{tabular}
  \caption{Comparison of English medical MCQA datasets}
  \label{tab:qa_comparison}
\end{table}



\subsection{Retrieval Augmentation Generation}


The basic RAG framework~\cite{lewis2020retrieval} couples a seq2seq model with a dense Wikipedia retriever, and has since been improved via query rewriting~\cite{chan2024rq}, entity graphs~\cite{edge2024local}, and document-structure–aware retrieval~\cite{li2025structrag}, though these methods mainly target general-domain text. For the medical domain, MedRAG~\cite{xiong-etal-2024-benchmarking} proposes a RAG pipeline with hybrid sparse–dense retrieval on MedCorps, and i-MedRAG~\cite{xiong2024improving} extends it with follow-up question generation and interactive reasoning. Self-BioRAG~\cite{jeong2024improving} adapts Self-RAG~\cite{asai2023self} with domain-specific retrieval triggers, while RAG\textsuperscript{2}~\cite{sohn2024rationale} leverages rationale-based queries and filtering. EXPRAG~\cite{ou2025experience} retrieves similar patient cases, and ClinicalRAG~\cite{lu2024clinicalrag} exploits medical entities to query corpora. However, existing medical RAG systems largely ignore question-specific expertise (e.g., subject area or certification) as explicit signals to guide retrieval and reasoning. Unlike Metadata-RAG~\cite{VanOudenhove2024SelfQueryRAG}, which parses metadata directly from the query, we inject expertise attributes inferred from the question by a model trained on our Q\&A dataset.

\section{EMSQA}
\subsection{Data Collection and Preprocessing}
We collected a total of 24.3K multiple-choice questions and their corresponding answer choices from practice tests available on 17 websites targeting the National Registry of Emergency Medical Technicians (NREMT) examination~\cite{nremt2025}. The NREMT exam is administered as a Computer Adaptive Test (CAT), which dynamically adjusts question difficulty based on the examinee’s performance, providing a personalized and efficient assessment. It certifies providers at 4 ascending levels of difficulty: entry‐level, Emergency Medical Responder (EMR); basic‐level, Emergency Medical Technician (EMT); intermediate‐level, Advanced EMT (AEMT); and advanced‐level, Paramedic. Our dataset covers all four certification levels. These practice tests assess examinees’ ability to apply medical knowledge, concepts, and principles, as well as their capacity to demonstrate fundamental patient-centered skills. 

The overall EMSQA dataset comprises questions from both public and private (subscription-based) websites, but we only release the portion derived from public materials because of copyright restriction of private websites (See \textbf{Appendix A.2 in Extended version} for more details).
To ensure the data quality, we did both automatic preprocessing and manual verification of the raw data as shown in Figure~\ref{fig:Overall approach}:
\begin{itemize}
    \item Heuristic rules were used to extract and remove special tokens like HTML tags, special symbols.
    \item Each question is well-structured and represented as a dictionary containing the following fields: the question text, answer options, correct answer, explanation, source URL, certification level, and subject area.
    \item Questions related to images and tables were excluded. All the questions are answerable using text inputs only. 
    \item All duplicate questions were removed by computing the Levenshtein distance between each pair of questions in the dataset. Pairs with a similarity score greater than 0.9 were considered duplicates and subsequently removed.
    \item All questions are manually labeled with subject areas, and both questions and answer choices have undergone human proofreading to correct any grammatical errors. A sample of 100 questions and KB documents was further verified by an EMT expert (see Appendix A.5).

\end{itemize}

\subsection{Data Statistics}
As shown in Table~\ref{tab:KB_PC_EMSQA_stats}, the dataset includes a total of 18,602 and 5,669 practice questions from public and private sources, respectively. We use the private questions exclusively for testing and split the public questions into train, validation, and test sets of 13,021, 1,860, 3,721 questions, with average token lengths of 18.27, 19.12, 18.99, respectively. More detailed statistics are provided in Appendix A.2.2.


\begin{table}[t!]
\centering
\small
\setlength{\tabcolsep}{1mm}
\begin{tabular}{ll|llcc}
\toprule
\textbf{Set} & \textbf{Size} & \textbf{Type} & \textbf{Criteria} 
  & \textbf{vs KB} & \textbf{vs PR} \\
\midrule
\multirow{3}{*}{Public} 
    & Train(13,021) & Semantic & Avg Sim & 79.21 & 66.45 \\
    & Val(1,860)    & \multirow{3}{*}{\makecell[l]{Syntactic\\(hit rate)}} & Vocab        & 82.95 & 21.14 \\
    & Test(3,721)   &          & Cpt w/o norm   & 41.65 & 8.87 \\
    &                &          & Cpt w/ norm    & 63.30 & 15.28 \\
\midrule
\multirow{1}{*}{Private} 
    & Test(5,669)   & Semantic & Avg Sim & 80.75 & 75.35 \\
    &                & \multirow{3}{*}{\makecell[l]{Syntactic\\(hit rate)}} & Vocab        & 90.89 & 28.26 \\
    &                &          & Cpt w/o norm   & 53.18 & 14.36 \\
    &                &          & Cpt w/ norm    & 72.49 & 22.66 \\
\bottomrule
\end{tabular}
\caption{Statistics by split for Public and Private EMSQA and Semantic and syntactic evaluation of QA overlap vs.\ KB/PR. Cpt: Concept; norm: medical normalization.}
\label{tab:KB_PC_EMSQA_stats}
\end{table}

\begin{figure}[t!]
  \centering
  \includegraphics[width=\columnwidth]{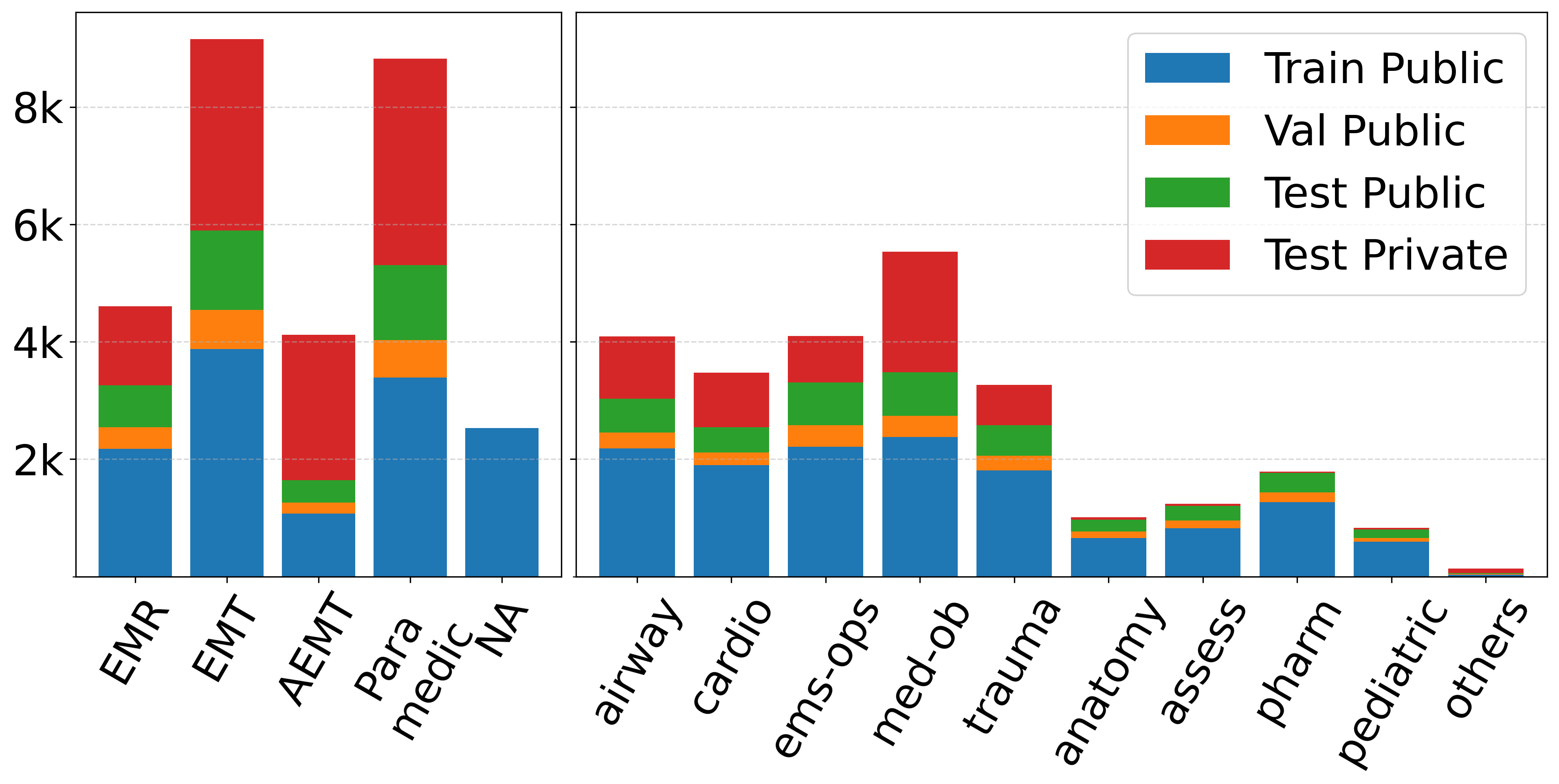}
  \caption{Distributions of Questions by Certification Level (Left) and Subject Area (Right).}
  \label{fig:statistics}
\end{figure}

Figure \ref{fig:statistics} details the distribution of questions by certification level and subject area. Because certification level is used for evaluation, all questions whose certification level marked with ``NA" are relegated to the training split, leaving the validation and test splits to include only questions with explicit EMS certification levels. Subject areas including ``airway", ``cardiology", ``EMS operations", ``medical\&OB", and ``trauma" dominate the corpus, which mirrors the five domains mandated by the NREMT examination guidelines. The remaining subject areas—``anatomy", ``assessment", ``pharmacology", ``pediatrics", and ``others"—are present as well, but they appear less frequently, reflecting their secondary coverage in the practice materials we collected.

\begin{figure*}[t!]
  \centering
  \includegraphics[width=\linewidth]{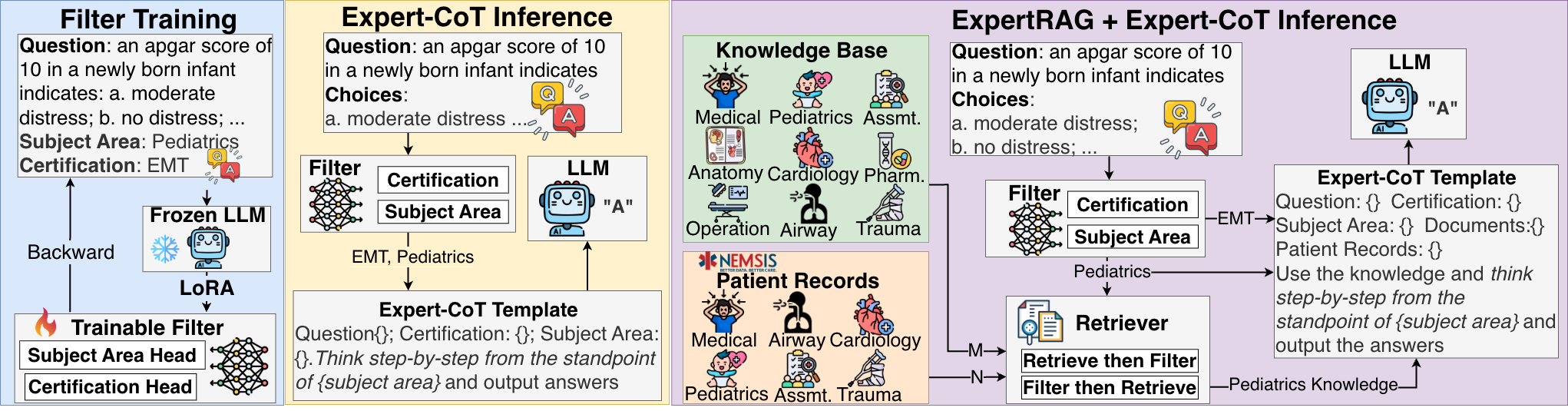}
  \caption{Expertise-Guided LLM Framework: Filter training, Expert-CoT Inference, and ExpertRAG Inference.
  }
  \label{fig:ExpertRAG}
\end{figure*}

\subsection{Knowledge Collection and Preprocessing}
To build an external KB for EMSQA, we curated 16 open-access EMS education resources from reputable websites. As shown in Figure~\ref{fig:Overall approach}, these resources span four media types: YouTube video transcripts~\cite{youtubeCD, youtubeEMP}, official EMS guidelines~\cite{odemsaDocs}, EMS education textbooks~\cite{redcrossEMR} or lecture slides~\cite{jbLearning}, and EMS flashcards~\cite{appEmtPrep}. We segmented each YouTube transcript into sections using the title cues supplied by the uploader. The PDF documents were converted to plain text with PyPDF2~\cite{pypdf2} and split to chapters using page ranges from their tables of contents. We then leveraged GPT-4o \cite{achiam2023gpt} to reorganize the raw chapter text into a coherent section-level hierarchy (See Appendix A.3.1). Finally, we manually audited each section to ensure its heading and text span aligned with the corresponding passage in the original PDF and corrected any discrepancies. Due to the lack of certification information in the sources, we only categorized the chapters into 10 subject areas based on their titles: ``airway\&ventilation", ``anatomy", ``assessment", ``cardiovascular", ``ems operations", ``medical\&ob", ``pediatrics", ``pharmacology", ``trauma" and ``others". Our final EMS KB comprises 39,652 sections, totaling 2,545,192 tokens and 34,110 unique vocabularies. 

To evaluate the helpfulness of the KB for EMSQA, we assess our KB’s coverage from both syntactic and semantic aspects. For semantic evaluation, we leverage MedCPT~\cite{jin2023medcpt} to retrieve, for each question, its most similar document from the KB and report the average similarity score between each question and its retrieved document. For syntactic evaluation, we removed stop words and computed vocabulary hit rate (KB $\cap$ QA / QA) between the KB and EMSQA. We also follow the methods in ~\cite{ge2024dkec} to extract EMS-related concepts defined in~\cite{preum2020emscontext}, both before and after UMLS normalization~\cite{bodenreider2004unified}. Table~\ref{tab:KB_PC_EMSQA_stats} reports both syntactic and semantic overlaps with our KB. Syntactic hit rates span from 41.65\% to 90.89\%. Semantic similarity, measured by average cosine similarity, was 79.21\% on public and 80.75\% on private data. This indicates that our KB can support answering most EMSQA questions. Appendix A.3 provides comprehensive details on web crawling, EMS concept extraction, KB statistics, and the overlap between EMSQA and the KB.

\subsection{Patient Care Report Collection and Preprocesing}
We used the National EMS Information System (NEMSIS)~\cite{dawson2006national} 2021 public research dataset as our source of patient records (PR). NEMSIS is a large tabular corpus in which each record includes fields such as dispatch details, scene information, initial assessment, EMS protocols and triage, vital signs, medications and procedural interventions, and patient history. As shown in Figure~\ref{fig:Overall approach}, we removed any field whose value was “NA,” “Not Applicable,” “Not Recorded,” or “Unknown.” If more than 30\% of a record’s fields were discarded, we excluded the entire record. Finally, we converted each remaining record into plain text by concatenating its key–value pairs and categorized the patient records into 6 subject areas based on their protocol field:  ``airway", ``assessment", ``cardiovascular", ``medical\&ob", ``pediatrics", and ``trauma". Our final corpus comprises 4,003,430 records with an average token length of 311.7. 
As shown in Table~\ref{tab:KB_PC_EMSQA_stats}, semantic coverage of NEMSIS over EMSQA is high, with patient records reaching 66.45\%/75.35\% similarity on the public/private split. In contrast, syntactic concept hit rates are much lower (8.87\%–28.26\%) than those of the KB. Extracted NEMSIS fields and summary statistics are provided in Appendix A.4.



\section{Methodology}
\subsection{Task Formulation}



Given the $i$th question $q_i$ and its answer options $\mathcal{O}_i = \{o_1, \dots, o_m\}$, the goal of MCQA is to maximize the likelihood of selecting the correct answer $a_i^* \in \mathcal{O}_i$. Let $\mathcal{R}$ be a retriever that takes $q_i$ as input and returns a set of relevant documents. A language model $f$ selects an answer $a_i$ as:
\begin{equation}
    a_i = \arg\max_{o \in \mathcal{O}_i} f(o \mid q_i, \mathcal{O}_i, \mathcal{R}(q_i))
\end{equation}

We propose an \textit{expertise-guided} LLM framework (see Figure~\ref{fig:ExpertRAG}) with an expertise classification module (called \textbf{Filter}), which infers the domain expertise attributes of subject area $s_i$ and certification level $l_i$ from the input question $q_i$, and incorporates them into $f$ using two strategies: (1) \textbf{Expert-CoT}, a prompting strategy that encodes $s_i$ and $l_i$ into a prompt template to guide $f$ based on question-specific expertise; and (2) \textbf{ExpertRAG}, a subject-area-specific $\mathcal{R}$ that retrieves  knowledge sources conditioned on $s_i$. 




\subsection{Filter Design}
To guide the LLM reasoning and RAG retrieval based on question-specific expertise, we train a lightweight LLM-based filter to infer the key expertise attributes, including question’s subject area and certification level. As shown in Figure~\ref{fig:ExpertRAG} (Left), we adopt LoRA~\cite{hu2022lora} to inject a small set of trainable parameters into the model while keeping the full LLM weights fixed. We augment the LoRA modules with two classification heads, $W_{\mathrm{sub}}$ and $W_{\mathrm{lvl}}$, that predict the question’s subject area and certification level, and optimize them jointly in a multi‐task setting~\cite{li2025mmsenseadaptingvisionbasedfoundation}. We append a special token \texttt{<classify>} at the end of each query, and extract the hidden state $h_i$ of this final token from the last layer and feed it into our two classification heads:
\begin{equation}
\begin{aligned}
h_i &= \mathrm{LM}_{\mathrm{last}}(q_i, \mathcal{O}_i \| \langle\mathrm{classify}\rangle), \\
(p_i^{\mathrm{sub}},\, p_i^{\mathrm{lvl}}) &= \big(\sigma(W_{\mathrm{sub}}^{\!\top} h_i),\, \sigma(W_{\mathrm{lvl}}^{\!\top} h_i)\big)
\end{aligned}
\end{equation}

where $\sigma$ is sigmoid function. We use binary cross‐entropy for subject area classification and cross‐entropy for certification classification. We set hyper-parameters $w_{\mathrm{sub}}$ and $w_{\mathrm{lvl}}$ to balance the multi-task loss. The overall training loss is:
\begin{equation}
\mathcal{L} = w_{\mathrm{sub}} \cdot \mathrm{BCE}\left({p}_i^{\mathrm{sub}},{y}_i^{\mathrm{sub}}\right) + w_{\mathrm{lvl}} \cdot \mathrm{CE}\left(p_i^{\mathrm{lvl}}, y_i^{\mathrm{lvl}}\right)
\end{equation}

During inference, given a question $q_i$ with answer options $\mathcal{O}_i$, the filter first predicts a certification-level probability distribution $p_i^{\mathrm{lvl}}$ and a multi-label subject area probability vector $p_i^{\mathrm{sub}}$. The predicted certification level and subject area set are computed as:
\begin{equation}
\hat{s}_i = \mathbf{1}\{p_i^{\mathrm{sub}} > 0.5\},
\quad
\hat{l}_i = \arg\max p_i^{\mathrm{lvl}}.
\end{equation}

\subsection{Expertise-Guided Prompting (Expert-CoT)}
Figure~\ref{fig:ExpertRAG} (Middle) illustrates our Expert-CoT prompting method. Standard CoT prompts encourage LLMs to reason step by step but do not specify where to begin. In contrast, Expert-CoT prompting guides the model’s reasoning by explicitly providing the subject area and certification level as starting point for the thought process. The final answer is generated by passing the predicted subject area $\hat{s}_i$ and certification level $\hat{l}_i$, into the Expert-CoT prompt template:
\begin{equation}
\hat{A}_i
  = f^{\mathrm{CoT\text{-}Expert}}\bigl(q_i,\;\mathcal{O}_i,\;\hat{l}_i,\;\hat{s}_i\bigr).
\end{equation}


\subsection{Expertise-Guided RAG (ExpertRAG)}

As shown in Figure~\ref{fig:ExpertRAG} (Right), for ExpertRAG the filter's predicted subject area guides the retriever to search for relevant knowledge base entries and patient records tailored to the question's subject area. The LLM then conditions on the predicted expertise and the retrieved documents to generate the final answer. Based on $q_i$ and the predicted subject area $\hat{s}_i$, we explore three retrieval strategies for retriever $\mathcal{R}$:

\begin{itemize}
    \item \textbf{Global}: Retrieve the top $M$ and $N$ evidence documents from the entire KB and PR, respectively. This serves as a baseline corresponding to standard RAG without any subject area filtering.
    \item \textbf{Filter then Retrieve (FTR)}: First filter the whole KB and PR to retain only documents matching the predicted subject area $\hat{s}_i$, then retrieve the top $M$ and $N$ documents from these filtered subsets.
    \item \textbf{Retrieve then Filter (RTF)}: First retrieve a larger candidate set from the whole KB and PR (e.g., $10 \times M$ from KB and $10 \times N$ from PR), then filter out documents whose subject area do not match $\hat{s}_i$, retaining the top $M$ and $N$ relevant documents.
\end{itemize}

The final answer is generated by passing the retrieved documents, along with the predicted subject area and certification level, into the RAG prompt template:
\begin{equation}
\hat{A}_i
  = f^{\mathrm{RAG}}\bigl(q_i,\;\mathcal{O}_i,\;\mathcal{R}(q_i, \hat{s}_i),\;\hat{l}_i,\;\hat{s}_i\bigr).
\end{equation}

\section{Experiments}
We conduct extensive experiments to evaluate
Expert-CoT and ExpertRAG methods by applying them to different baseline LLMs and comparing their performance to SOTA LLMs and RAGs. We aim to answer three research
questions:\\
\textbf{RQ1:} Where do SOTA LLMs shine or stumble on EMSQA across subject areas and certification levels?\\
\textbf{RQ2:} How much does explicit expertise injected by Expert-CoT and ExpertRAG lift baseline accuracy?\\
\textbf{RQ3:} Can expertise-aware LLMs pass the NREMT standardized tests  at different certification levels?

\subsection{LLM Baselines}
We use three categories of SOTA baseline models: 
(1) \textbf{Open-source LLMs}: we select Qwen3-32B~\cite{qwen3technicalreport}, and LLama-3.3-70B~\cite{grattafiori2024llama} because of their great performance in multiple domains;
(2) \textbf{Medical LLMs}: we select OpenBioLLM-70B~\cite{OpenBioLLMs}, which currently leads the Open Medical-LLM Leaderboard~\cite{open-medical-llm-leaderboard}. 
(3) \textbf{Closed-source LLMs}: we choose OpenAI-o3~\cite{brown2020language} and Gemini-2.5-pro~\cite{team2023gemini}, both of which achieve top results on a range of benchmarks. We further apply 0- to 64-shot, CoT~\cite{wei2022chain}, and Expert-CoT prompting to each baseline to benchmark the performance across prompt strategies. 
\subsection{RAG Baselines}
We select the following SOTA medical RAG models due to their superior performance and code availability: (1) MedRAG~\cite{xiong-etal-2024-benchmarking}, which is a RAG toolkit combining multiple medical documents; (2) i-MedRAG~\cite{xiong2024improving}, which iteratively refines medical queries via multi-step retrieval; (3) Self-BioRAG~\cite{jeong2024improving}, which self-reflectively decides when to retrieve biomedical texts and then generates the answer; (4) Qwen3-4B + KB, which is a vanilla RAG pipeline with our collected KB as retrieval corpora; (5) Qwen3-4B + PR, a vanilla RAG pipeline with our collected PR as retrieval corpora; (6) Qwen3-4B + Global, a vanilla RAG pipeline with  both KB and PR as retrieval corpora. We also include (7) Qwen3-4B with 0-shot and (8) Qwen3-4B with CoT prompting as baselines. For a fair comparison, we applied RAG with Qwen3-4B across all methods, except Self-BioRAG, which is trained from scratch. All baseline RAG methods used CoT prompting.

\subsection{Implementation Details}
For Expert-RAG, we use Qwen3 as the core LLM, as it is the best-performing open-source model in our benchmarking.
We employ MedCPT as our retriever due to its strong performance in medical domain and widespread use in SOTA RAG. We fix the number of retrieved documents at $M=32$ for KB retrieval and $N=8$ for PR retrieval. All KB and PR documents are chunked with a window of 512 tokens with an overlap of 128 tokens. Since some questions in EMSQA have multiple correct answers, we report both exact-match accuracy (Acc) and sample-based F1~\cite{khashabi2018looking}. 

 ßTo train the filter, we fine-tune LoRA modules with rank $r{=}8$, scaling factor $\alpha{=}16$, and a dropout rate of 0.05, using the sequence length of 128 tokens. To balance the certification and subject area classification objectives, we apply DWA~\cite{liu2019end} with $T=2$ to dynamically adjust $w_{\mathrm{cat}}$ and $w_{\mathrm{lvl}}$. We use AdamW~\cite{loshchilov2017decoupled} optimizer and regularization with a weight decay of 0.01. We set the decision threshold of 0.5 for subject area classification. We fix the random seed at 42, and run all experiments on NVIDIA H200 GPUs. 

\begin{figure}[t!]
  \centering
  \includegraphics[width=\linewidth]{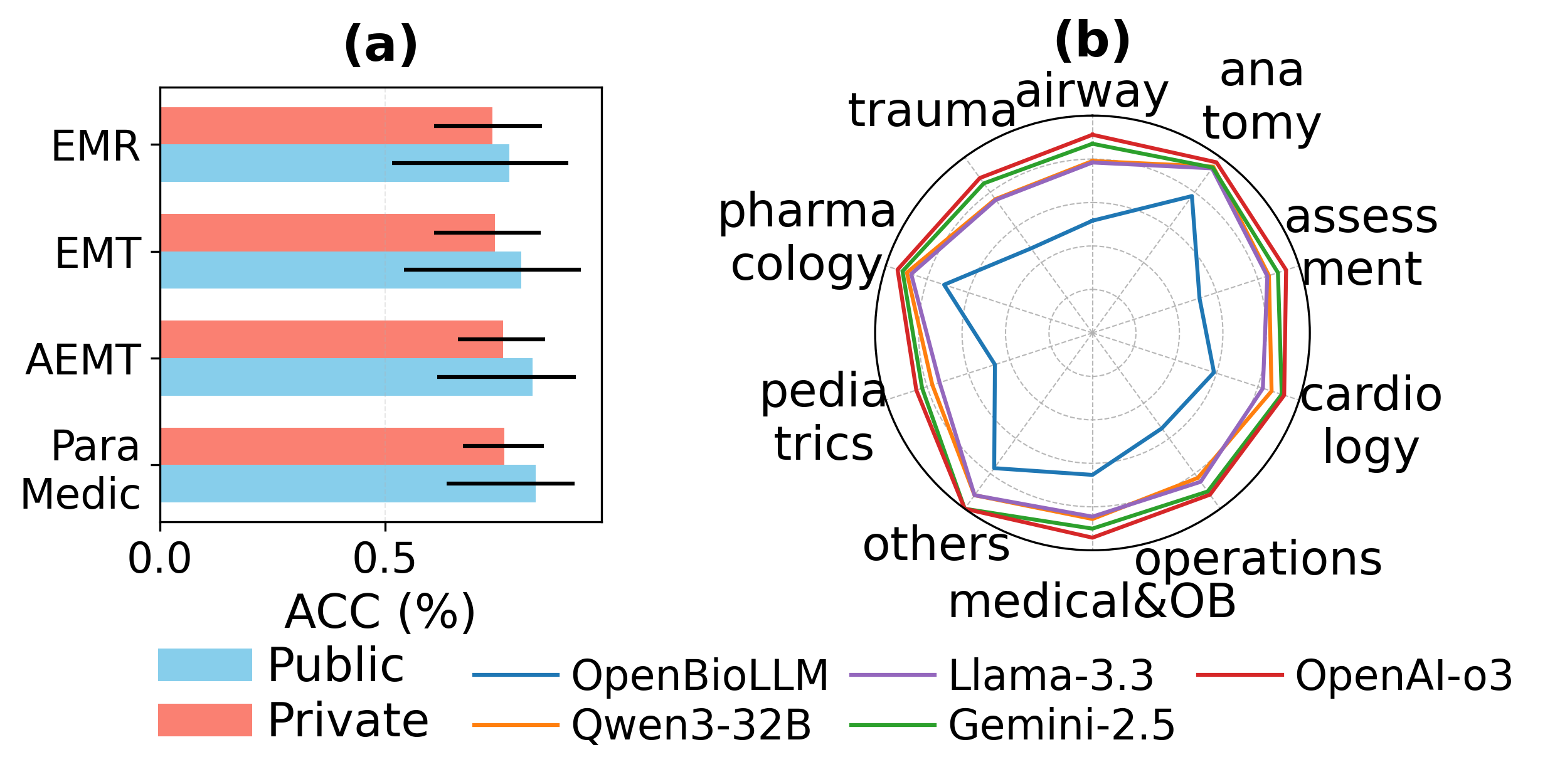}
  \caption{%
    \textbf{(a)} Certification level 0-shot performance. \textbf{(b)} Subject area 0-shot performance on the Public dataset 
    }
  \label{fig:category_certification_results}
\end{figure}

\begin{table}[t]
  \centering
  \setlength{\tabcolsep}{1mm}
  \small
  \begin{tabular}{l l  c c  c c}
    \toprule
    \textbf{Model}      & \textbf{Prompt}              & \multicolumn{2}{c}{\textbf{Public}}    & \multicolumn{2}{c}{\textbf{Private}}   \\
    \cmidrule(lr){3-4} \cmidrule(lr){5-6}
               &                    & Acc & F1           & Acc & F1           \\
    \midrule
    \multirow{3}{*}{OpenBioLLM} 
            & 0-shot             & 57.67   & 57.76      & 63.86   & 64.76      \\
            & CoT             & 59.88    & 60.34     & 67.01  & 67.77    \\
            \multirow{1}{*}{\quad(GT)} & Expert-CoT             & \textbf{61.92}   & \textbf{62.03}      & \textbf{68.75}   & \textbf{69.82}   \\
    \multirow{1}{*}{\quad(Filter)} & Expert-CoT  &  61.32  & 61.93     & 67.79  &  68.32     \\
    \midrule
    \multirow{3}{*}{Llama-3.3}
            & 0-shot             & 81.69   & 82.69      & 78.06   & 78.77      \\
            & CoT             & 81.89   & 83.08      &  85.16  & 86.35      \\
            \multirow{1}{*}{\quad(GT)} & Expert-CoT             & \textbf{82.42}   & \textbf{83.35}      & 86.49   & 87.62      \\
    \multirow{1}{*}{\quad(Filter)} & Expert-CoT      & 82.40   & 83.18      & \textbf{86.63}   & \textbf{87.65}      \\
    \midrule
    \multirow{8}{*}{Qwen3-32B}
               & 0-shot             & 83.55   & 83.55      & 85.11   & 85.89      \\
               & 4-shot             & 84.41   & 84.41      & 85.48   & 86.13      \\
               & 32-shot            & 81.13   & 81.13      & 82.22   & 83.41      \\
               & 64-shot            & 82.48   & 82.48      & 86.22   & 87.26      \\
               & CoT   & 84.96   & 84.97      & 88.78   & 90.13      \\
               \multirow{1}{*}{\quad(GT)} & Expert-CoT      & \textbf{85.70}   & \textbf{85.71}      & \textbf{89.73}   & 90.98  \\
    \multirow{1}{*}{\quad(Filter)} & Expert-CoT      & 85.57   & 85.60      & 89.50   & \textbf{91.20}      \\
    \midrule
    OpenAI-o3  & 0-shot             & \textbf{92.39}   & \textbf{92.39}      & --        & --           \\
    Gemini-2.5 & 0-shot         & 89.36   & 89.36      & --        & --           \\
    \bottomrule
  \end{tabular}
  \caption{Accuracy and F1 (\%) of LLMs under Public vs.\ Private Data. GT/Filter: Ground-truth/predicted expertise.
  }
  \label{tab:benchmark_llm_results}
\end{table}

\section{Experimental Results}



\subsection{LLM Benchmarking}
To evaluate the overall performance of SOTA LLMs on EMSQA (\textbf{RQ1}), we benchmark multiple LLMs under different prompting strategies.  
Figure~\ref{fig:category_certification_results} and Table~\ref{tab:benchmark_llm_results} present the results. We highlight several key findings from the evaluation:

\begin{table}[t!]
\centering
\small
\setlength{\tabcolsep}{1mm}
\begin{tabular}{@{}l l l cc cc@{}}
\toprule
\textbf{Split} & \textbf{Model} & \textbf{Method} & \multicolumn{2}{c}{\textbf{Subject Area}} & \multicolumn{2}{c}{\textbf{Certification}} \\
               &                &                      & miF & maF & miF & maF \\
\midrule
\multirow{4}{*}{Public} 
       & Filter      & LoRA     & \textbf{80.72} & \textbf{71.92} & \textbf{65.87} & \textbf{63.45} \\
       & Qwen3-4B    & 0-shot   & 55.43      & 51.61      & 45.77      & 30.49      \\
       & Qwen3-4B    & 4-shot   & 56.33      & 54.42      & 45.46      & 30.28      \\
       & Qwen3-4B    & CoT      & 59.72  & 55.66      & 47.80      &  35.77     \\
\midrule
\multirow{4}{*}{Private} 
       & Filter      & LoRA     & \textbf{79.06} & \textbf{70.48} & \textbf{65.54} & \textbf{63.50} \\
       & Qwen3-4B    & 0-shot   & 42.93      & 31.73      & 44.12      & 25.01       \\
       & Qwen3-4B    & 4-shot   & 45.76      &  34.08     & 44.04      & 29.41      \\
       & Qwen3-4B    & CoT      & 46.22      & 35.49       & 47.70       & 31.92      \\
\bottomrule
\end{tabular}
\caption{Expertise Classification Performance}
\label{tab:filter_result}
\end{table}

\newcommand{\dAF}{\Delta_{\mathrm{Acc/F1}}}

\begin{table}[t!]
\centering
\small
\setlength{\tabcolsep}{1.5mm}
\begin{tabular}{@{}l l cc cc@{}}
\toprule
\textbf{Model}      & \textbf{Description}   & \multicolumn{2}{c}{\textbf{Public}} & \multicolumn{2}{c}{\textbf{Private}} \\
                    &                        & Acc          & F1            & Acc           & F1             \\
\midrule

\multicolumn{6}{@{}l}{\textbf{No-RAG Baselines}}\\
Qwen3-4B            & 0-shot                 & 70.99        & 71.01         & 69.88         & 69.95          \\
Qwen3-4B            & CoT                    & 72.35        & 73.09         & 70.58         & 72.02          \\
\midrule
\multicolumn{6}{@{}l}{\textbf{RAG Baselines + CoT}}\\
MedRAG              & RAG on Med             & 74.31        & 74.41         & 71.12         & 73.33          \\
i-MedRAG            & Iterative RAG          & 77.96 & 78.00    & 74.02         & 76.35          \\
Self-BioRAG         & SelfRAG on Bio         & 55.71        & 58.84         & 45.72         & 49.67          \\
Qwen3-4B            & KB                     & 76.49        & 76.07         & 75.02         & 76.53          \\
Qwen3-4B            & PR                     & 73.02        & 73.96         & 70.54         & 72.38          \\
Qwen3-4B            & Global                 & \underline{78.12}        & \underline{79.17} & \underline{75.46} & \underline{76.87} \\
\midrule
\multicolumn{6}{@{}l}{\textbf{RAG Baselines + Expert-CoT}\hfill$\dAF=\mathord{+}1.38/\mathord{+}0.46$} \\
Qwen3-4B           & KB      & 78.02        & 79.04         & 76.01         & 76.25         \\
Qwen3-4B           & PR      & 73.82        & 73.82         & 71.53         & 72.96         \\
Qwen3-4B           & Global  & \textbf{79.59}        & \textbf{79.61}         & \textbf{76.75}         & \textbf{77.35}         \\
\midrule
\multicolumn{6}{@{}l}{\textbf{ExpertRAG-GT + CoT}\hfill$\dAF=\mathord{+}3.35/\mathord{+}2.71$} \\
ExpertRAG           & FTR    & 80.97        & 81.34         & 79.13         & 80.00         \\
ExpertRAG           & RTF    & \textbf{81.11}        & \textbf{81.45}         & \textbf{79.17}         & \textbf{80.01}         \\
\midrule

\multicolumn{6}{@{}l}{\textbf{ExpertRAG-GT + Expert-CoT}\hfill $\dAF=\mathord{+}4.59/\mathord{+}3.69$} \\  
ExpertRAG           & FTR    & 81.62        & 81.65         & 80.40         & 81.02         \\
ExpertRAG           & RTF    & \textbf{82.24} & \textbf{82.26} & \textbf{80.51} & \textbf{81.16} \\
\midrule
\multicolumn{6}{@{}l}{\textbf{ExpertRAG-Filter + Expert-CoT}\hfill $\dAF=\mathord{+}3.44/\mathord{+}2.59$} \\  

ExpertRAG           & FTR    & \textbf{80.99}        & \textbf{80.99}         & 79.45         & 80.16         \\
ExpertRAG           & RTF    & 80.95        & 80.96         & \textbf{79.47}         & \textbf{80.22}         \\
\bottomrule
\end{tabular}
\caption{End-to-end RAG Performance and Ablation Study on ExpertRAG and Expert-CoT.
}
\label{tab:ExpertRAG_results}
\end{table}

\begin{table*}[t!]
  \centering
  \small
  \setlength{\tabcolsep}{1mm}
  \begin{tabular*}{\textwidth}{@{\extracolsep{\fill}} l l  c c c c | c c c c | c c c c | c c c c }
    \toprule
    \textbf{Model} & \textbf{Description}
      & \multicolumn{4}{c}{\textbf{EMR}}
      & \multicolumn{4}{c}{\textbf{EMT}}
      & \multicolumn{4}{c}{\textbf{AEMT}}
      & \multicolumn{4}{c}{\textbf{Paramedic}} \\
    \cmidrule(lr){3-6}
    \cmidrule(lr){7-10}
    \cmidrule(lr){11-14}
    \cmidrule(lr){15-18}
    & & Pass & Score & Acc & T
      & Pass & Score & Acc & T
      & Pass & Score & Acc & T
      & Pass & Score & Acc & T \\
    \midrule
    \multirow{2}{*}{Qwen3-4B}
      & 0-shot 
        & \ding{55} & 809 & 64.18 &  33
        & \ding{55} & 940 & 74.07 &  33
        & \ding{55} & 940 & 71.64 &  33
        & \ding{55} & 940 & 71.72 & 35 \\
      & Expert-CoT 
        & \ding{55} & 940 & 72.42 &  59
        & \ding{55} & 940 & 76.73 & 73 
        & \ding{51} & 1179 & 80.41 & 76 
        & \ding{55} & 940 & 76.03 & 87 \\
    \midrule
    \multirow{2}{*}{ExpertRAG-4B}
      & FTR+Expert-CoT
        & \ding{51} & 1218 & 84.21 & 66
        & \ding{55} & 940 & 78.76 & 61
        & \ding{55} & 940 & 77.31 & 69
        & \ding{55} & 940 & 79.67 & 74\\
      & RTF+Expert-CoT
        & \ding{55} & 940 & 76.47 & 59
        & \ding{51} & 1185 & 81.30 & 93
        & \ding{51} & 1190 & 83.53 & 67
        & \ding{55} & 940 & 77.61 & 86\\

    \midrule
    \multirow{2}{*}{Qwen3-32B} 
      & 0-shot
        & \ding{51} & 1207 & 82.65 & 22
        & \ding{51} & 1140 & 81.63 & 23
        & \ding{51} & 1280 & 85.92 & 26
        & \ding{51} & 1163 & 81.58 & 29\\
      & Expert-CoT
        & \ding{51} & 1261 & 86.27 & 50
        & \ding{51} & 1255 & 86.96 & 52
        & \ding{51} & \underline{1310} & \underline{89.11} & 61
        & \ding{51} & \textbf{1292} & \textbf{89.01} &  57\\
    \midrule
    \multirow{2}{*}{ExpertRAG-32B}
      & FTR+Expert-CoT
        & \ding{51} & \textbf{1350} & \textbf{92.22} & 75
        & \ding{51} & \underline{1292} & \underline{89.01} & 76
        & \ding{51} & 1215 & 84.60 & 86
        & \ding{51} & 1228 & 83.93 & 125\\
      & RTF+Expert-CoT
        & \ding{51} & \textbf{1350} & \textbf{92.22} & 75
        & \ding{51} & \textbf{1328} & \textbf{92.32} & 82
        & \ding{51} & \textbf{1356} & \textbf{92.31} & 82
        & \ding{51} & \underline{1276} & \underline{88.04} & 99 \\
    \bottomrule
  \end{tabular*}
  \caption{Pass (\ding{51}) or Fail (\ding{55}) Summary of Models by Simulation Certification Test. T: Overall Time (min).}
  \label{tab:cat-test}
\end{table*}

\textbf{Closed-source models outperform open-source models.} In particular, OpenAI-o3 consistently achieves the highest overall accuracy of 92.39. Among open-source models, Qwen3-32B achieves the best accuracy of 85.70, though a significant gap remains compared to closed-source models.

\textbf{Few-shot prompting improves accuracy up to a point.} We varied the number of in‑context exemplars from 0 to 64 (See Appendix A.7.2) and observed that incorporating a small number of examples yields substantial gain over the zero‑shot baseline. However, adding examples beyond a certain point leads to diminishing or no improvement.

\textbf{Baseline LLMs underperform on easier questions.} As shown in Figure~\ref{fig:category_certification_results}a, across all models, we observe that performance is lowest for the EMR certification level (the most basic tier in the NREMT exam) and highest for the Paramedic level. This may be due to smaller data size for EMR level and the procedural nature of EMR questions, whereas Paramedic questions aligning more closely with the medical content seen during LLM pretraining.

\textbf{LLMs falter on the core NREMT domains}. Figure \ref{fig:category_certification_results}b shows that models reliably answer ``pharmacology'' and ``anatomy'' questions but struggle in ``pediatrics'' and core areas such as ``trauma'',``airway'', ``operations'', and ``cardiology''.
One possible reason is subject area complexity. 
Questions in the former areas often need single-hop, fact-based queries solvable in a zero-shot manner, whereas the latter demand multi-hop reasoning and richer EMS knowledge.

\subsection{Expertise Classification Performance}
 The performance of our Filter vs. LLM baselines (0-shot, 4-shot, and CoT) for expertise classification are shown in Table~\ref{tab:filter_result}. Since subject area classification is a \textit{multi-label} task and certification classification is a \textit{multi-class} task, we report micro f1-score (miF) and macro f1-score (maF). Results show our Filter trained with LoRA with two classification heads significantly outperforms the baseline LLMs.

\subsection{Expert-CoT Evaluation}
To assess how domain expertise, injected via Expert-CoT, influences reasoning (\textbf{RQ2}), we compare Expert-CoT with different prompting strategies under multiple LLMs. As shown in Table~\ref{tab:benchmark_llm_results},
\textbf{Expert-CoT help guide LLM reasoning.} Integrating domain expert knowledge via CoT‑Expert guides reasoning towards appropriate context and consistently boosts CoT prompting performance by up to 2.05\% across models. Also, using the predicted expertise attributes (Filter) vs. the ground-truth attribute annotations (GT) yields comparable performance for Expert-CoT, demonstrating the Filter's strong performance, as also shown in Table~\ref{tab:filter_result}.

\subsection{Ablation Study on Expert-CoT and ExpertRAG}
We further evaluate the effect of injecting expertise into CoT and RAG~(\textbf{RQ2}) using an ablation study with six configurations, as shown in Table~\ref{tab:ExpertRAG_results}:
(1) \textit{No-RAG}, 
(2) \textit{RAG+CoT} with a standard global retriever on EMS PR and KB,
(3) \textit{RAG+Expert-CoT}, 
(4) \textit{ExpertRAG-GT+CoT}, 
(5) \textit{ExpertRAG-GT+Expert-CoT}, and 
(6) \textit{ExpertRAG-Filter+Expert-CoT}.
An ablation study on certification and subject area is presented in Appendix A.8. The best configuration outperforms the baseline by 4.59 / 3.69 points in \text{Acc} / \text{F1} (See error analysis in Appendix A.9).

\textbf{Effect of PR and KB.} (2) vs. (1) isolates the gain of adding PR and KB as retrieval documents with a standard global retriever. Results show KB brings more improvement than PR, and combing both yields the best performance.

\textbf{Effect of Expert-CoT.}
(3) vs. (2) (and (5) vs. (4)) ablates the additional gain from Expert-CoT on RAGs, showing that expertise-aware reasoning is better than standard reasoning.

\textbf{Effect of Expert-RAG.}
(4) vs. (2) (and (5) vs. (3)) ablate the impact of our Expert-RAG (FTR/RTF) compared to standard RAG with a global retriever. With ground-truth subject area and certification level, ExpertRAG consistently outperforms SOTA RAG baselines, highlighting the value of expertise-guided retrievers. Both FTR and RTF outperform global retrieval, with RTF achieving better performance.

\textbf{Effect of Filter.} (6) vs. (5) measures the effect of using the Filter's predicted expertise vs. ground-truth expertise annotations.  There is a small performance drop, but ExpertRAG-Filter still outperforms the best baseline.

\subsection{NREMT Computer Adaptive Simulation Tests}
To investigate whether our best models can be certified in NREMT exam (\textbf{RQ3}), we subscribed to MedicTests~\cite{medictests2025}, the NREMT Computer Adaptive Simulation Test. The simulation exam consists of 80-150 adaptively selected questions and must be completed within 2.5 hours. The NREMT cognitive exam is scored on a 100–1500 scale, with 950 as the passing threshold. 
We evaluated our Expert-CoT and ExpertRAG models, along with 0-shot baselines. The expertise-augmented models used the trained Filter to predict the subject area and certification. The Pass/Fail outcomes are shown in Table~\ref{tab:cat-test}. All models completed the exam within the allotted time, though expertise-augmented models took much longer. These models consistently achieved higher test scores and significantly improved accuracy relative to baseline LLMs. However, performance varied with model size. 4B models failed at one or more certification levels, whereas 32B models passed all four. ExpertRAG‑32B with the RTF retrieval strategy achieved the highest overall score across certifications. Notably, although the smaller LLM did not pass the test, it benefited the most from expertise augmentation, by showing the largest accuracy gains and achieving scores near or above the passing threshold.



\section{Conclusion}
This paper presents a domain expertise-aware LLM framework for medical multiple-choice question answering that infers and incorporates expertise to guide LLM reasoning and RAG retrieval. We introduce \textbf{EMSQA}, the first large-scale labeled MCQA dataset for EMS with subject area and certification-level annotations, along with curated EMS knowledge bases. We propose \textbf{Expert-CoT}, which guides LLM reasoning by injecting expertise attributes into prompts, and \textbf{ExpertRAG}, which retrieves expertise-specific knowledge for augmented generation. Experiments show that our expertise-aware prompting and RAG strategies significantly improve performance over baselines. Importantly, the expertise-augmented LLMs pass the NREMT simulation tests across all EMS certification levels. EMSQA provides a new benchmark for MCQA research in medical domain, and our proposed expertise-aware LLM framework can be applied to other medical MCQA datasets with similar or other expertise attributes. 

\section{Ethics Statement}
All models studied in this work are research prototypes and not approved medical devices. They must not be used as the sole basis for diagnosis or treatment decisions. Outputs should serve only as a reference for licensed healthcare professionals, who remain fully responsible for clinical judgment and patient care. The models may generate incorrect, incomplete, or biased recommendations and may not reflect up-to-date guidelines. All experiments were conducted in simulation, with no model outputs used to influence real-world patient care. All private data were kept confidential.

\section*{Acknowledgments}
This work was supported by the award 70NANB21H029 from the U.S. Department of Commerce, National Institute of Standards and Technology (NIST), and a research grant from the Commonwealth Cyber Initiative (CCI).

\bibliography{aaai2026}

\section{Appendix}
\section{A.1\quad Introduction}
This is the technical appendix for the paper "Expert-Guided Prompting and Retrieval-Augmented Generation for
Emergency Medical Service Question Answering". 
\begin{itemize}
  \item \textbf{A.2 EMSQA Data} \dotfill \pageref{sec:2}
  \begin{itemize}
      \item \textit{A.2.1 EMSQA Data Collection} \dotfill \pageref{sec:2.1}
      \item \textit{A.2.2 EMSQA Data Statistics} \dotfill \pageref{sec:2.2}
  \end{itemize}
  \item \textbf{A.3 Knowledge Base Preprocessing \& Statistics}   \dotfill \pageref{sec:3}
  \begin{itemize}
    \item \textit{A.3.1 Prompt for Organizing Chapter Text} \dotfill \pageref{sec:3.1}
    \item \textit{A.3.2 Prompt for EMS Concept Extraction} \dotfill \pageref{sec:3.2}
    \item \textit{A.3.3 UMLS Concept Normalization} \dotfill \pageref{sec:3.3}
    \item \textit{A.3.4 Knowledge and EMSQA Overlap Statistics} \dotfill \pageref{sec:3.4}
    \item \textit{A.3.5 Knowledge Base Collection Details} \dotfill \pageref{sec:3.5}
  \end{itemize}
  
    \item \textbf{A.4 NEMSIS Patient Care Report Preprocessing \& Statistics} \dotfill \pageref{sec:4}
    \begin{itemize}
        \item \textit{A.4.1 NEMSIS Patient Record Preprocessing} \dotfill \pageref{sec:4.1}
        \item \textit{A.4.2 NEMSIS Patient Record Statistics} \dotfill \pageref{sec:4.2}
    \end{itemize}
    \item \textbf{A.5 Human Evaluation of EMSQA \& Knowledge Base} \dotfill \pageref{sec:5}
    \item \textbf{A.6 Retrieval Strategy Performance} \dotfill \pageref{sec:6}
    \item \textbf{A.7 Benchmark LLMs} \dotfill \pageref{sec:7}
    \begin{itemize}
        \item \textit{A.7.1 Zero-shot Performance per Subject Area} \dotfill \pageref{sec:7.1}
        \item \textit{A.7.2 Detailed statistics of Benchmarks} \dotfill \pageref{sec:7.2}
    \end{itemize}
    \item \textbf{A.8 Ablation Study on Certification Level and Subject Area} \dotfill \pageref{sec:8}
    \item \textbf{A.9 Error Analysis} \dotfill \pageref{sec:9}
    \begin{itemize}
        \item \textit{A.9.1  Error Types}\dotfill \pageref{sec:9.1}
        \item \textit{A.9.2 Error per Subject Area and Certification} \dotfill \pageref{sec:9.2}
    \end{itemize}

\end{itemize}


\begin{table*}[ht]
  \centering
  \begin{tabular}{@{}l c l c@{}}
    \toprule
    \textbf{Source} & \textbf{\# QAs} & \textbf{Certification Level(s)} & \textbf{Explanation} \\
    \midrule
    \multicolumn{4}{@{}l}{\textbf{Public}} \\
    \cite{emt5th} 
      & 955   & EMT                             & \checkmark \\
    \cite{nremtprak}     
      & 240   & EMT                             & \checkmark \\
    \cite{smartmedic}             
      & 537   & EMT                             & \ding{55}   \\
    \cite{unionprep}          
      & 150   & EMT                             & \checkmark \\
    \cite{montgomerymd}     
      & 130   & EMT                             & \checkmark \\
    \cite{mometrix}               
      & 138   & EMT, AEMT, Paramedic            & \checkmark \\
    \cite{practicetestgeeks}      
      &  99   & EMR                             & \checkmark \\
    \cite{mycprcert}  
      &  74   & EMR, EMT, AEMT                  & \ding{55}   \\
    \cite{quizizz}                
      & 1,554 & EMR, EMT, AEMT, Paramedic       & \ding{55}   \\
    \cite{minniquiz}             
      & 4,444 & EMR, EMT, Paramedic             & \ding{55}   \\
    \cite{quizlet}               
      & 9,084 & EMR, EMT, AEMT, Paramedic, NA   & \ding{55}   \\
    \cite{careeremployer}        
      & 100   & AEMT                            & \checkmark \\
    \cite{emtfourdummies}  
      & 456   & EMT                             & \checkmark \\
    \cite{learningexpresshub}    
      & 1,336 & Paramedic, EMT                  & \checkmark \\
    \cite{pocketprep}        
      &  47   & EMR, EMT, AEMT, Paramedic       & \checkmark \\
    \midrule
    \multicolumn{4}{@{}l}{\textbf{Private}} \\
    \cite{appEmtPrep}          
      & 4,843 & EMR, EMT, AEMT, Paramedic, Critical Care & \checkmark \\
    \cite{jbLearning}           
      & 1,285 & EMT                             & \checkmark \\
    \bottomrule
  \end{tabular}
  \caption{Summary of NREMT practice question sources in EMSQA.}
  \label{tab:emsqa_sources_details}
\end{table*}

\section{A.2\quad EMSQA Data}
\label{sec:2}

\subsection{A.2.1\quad EMSQA Data Collection}
\label{sec:2.1}

In Table~\ref{tab:emsqa_sources_details}, we present detailed information on 17 public and private EMSQA resources, including each source’s name and URL, number of questions, certification levels, and the availability of explanations.

\subsection{A.2.2\quad EMSQA Data Statistics}
\label{sec:2.2}
In Table~\ref{tab:split_stats}, we show the detailed EMSQA dataset statistics. The dataset includes a total of 18,602 and 5,669 practice questions from public and private sources, respectively. We use the private questions exclusively for testing and split the public questions into train, validation, and test sets of 13,021, 1,860, 3,721 questions, with average token lengths of 18.27, 19.12, 18.99, respectively.
\begin{table*}[htbp]
  \centering
  \setlength{\tabcolsep}{1mm}
  \begin{tabular}{@{} c l c r c c c r r @{}}
    \toprule
    \textbf{Data} 
      & \textbf{Split} 
      & \makecell[c]{\textbf{\#Explanations}}
      & \makecell[c]{\textbf{\#Choices}\\(avg/max)} 
      & \makecell[c]{\textbf{\#Answers}\\(avg/max)} 
      & \makecell[c]{\textbf{Question Tokens}\\(avg/max)} 
      & \makecell[c]{\textbf{Choice Tokens}\\(avg/max)} 
      & \makecell[c]{\textbf{Tokens}}
      & \makecell[c]{\textbf{Vocab}} \\
    \midrule
    \multirow{4}{*}{Public} 
      & Train (13,021) & 2217 & 4.01 / 7.00 & 1.00 / 3.00 & 18.27 / 218 & 6.28 / 240 & 565,303 & 14,017 \\
      & Val (1,860) & 383 & 3.99 / 5.00 & 1.00 / 3.00 & 19.12 / 155 & 6.01 / 44  &  80,215 &  6,629 \\
      & Test (3,721) & 773 & 4.01 / 6.00 & 1.00 / 3.00 & 18.99 / 135 & 6.10 / 60  & 161,464 &  8,913 \\
      & Total (18,602) & 3132 & 4.01 / 7.00 & 1.00 / 3.00 & 18.50 / 218 & 6.22 / 240 & 806,982 & 16,032 \\
    \midrule
    Private 
      & Test  (5,669) & 5451 & 4.01 / 6.00 & 1.06 / 4.00 & 30.44 / 355 & 5.46 / 47  & 296,673 & 10,637 \\
    \bottomrule
  \end{tabular}
  \caption{Statistics by split for Public and Private EMSQA
  }
  \label{tab:split_stats}
\end{table*}

\section{A.3\quad Knowledge Base Preprocessing \& Statistics}
\label{sec:3}

\begin{figure}[ht]
  \centering
  \begin{tcolorbox}[breakable=false, width=\columnwidth,]
\label{zero-shot}
\textbf{PROMPT:}\\
Well formated the unstructured text by the folllowing rules:\\
1. Fix awkward or broken line breaks within sentences or paragraphs.\\
2. Separate paragraphs with a single blank line.\\
3. Remove figure captions and references (e.g., “Figure 8–1”, “see Figure…”).\\
4. Remove page numbers (e.g., “239”).\\
5. Remove attribution or copyright lines, such as: {defined copyrights}\\
6. **Do not rephrase, reword, rewrite, or summarize; do not add any other words in your response**.\\
7. If the unstructured text contains subtitles, treat each subtitle as a key and store its corresponding paragraph(s) as the value. Return only the resulting JSON.\\
Here is the unstructured text to format: \textbf{RAW TEXT}
\end{tcolorbox}
  \caption{Prompt for organizing chapter text}
  \label{fig:organize_chapter}
\end{figure}

\subsection{A.3.1\quad Prompt for Organizing Chapter Text}
\label{sec:3.1}
As shown in Figure~\ref{fig:organize_chapter}, we present the prompt that instructs the GPT-4o to divide each chapter’s raw text into clearly labeled sections.
\label{prompt:ems_kb_clean}

\subsection{A.3.2\quad Prompt for EMS Concept Extraction}
\label{sec:3.2}
In main paper \textbf{Table 3: Semantic and syntactic evaluation of QA overlap vs. KB/PR}, we present the concept overlap in Syntactic metrics, we present the prompt we used for EMS concept extraction in Figure~\ref{fig:cpt_extraction}.

\begin{figure}[t!]
  \centering
  \begin{tcolorbox}[breakable=false,width=\columnwidth,]
\label{zero-shot}
\textbf{PROMPT:}\\
Extract all EMS concepts from the following text.\\
Here are some examples of EMS concepts: gelastic epilepsy, visual seizure, pallor, pale color, aox4, pare down, cut down\\
Return all the extracted EMS concepts as a lowercase letter in strict JSON format, like: ["fever", "cardiac arrest"]\\

Here is one example: \\
Text: Early symptoms include cough, wheezing, shortness of breath. Lung transplantation is an option.
Response: Let’s think step by step,\\

Step1: label the tokens one by one "EMS concept", or "none".\\
-Early: none\\
-symptoms: none\\
-include: none\\
-cough: EMS concept\\
-wheezing: EMS concept\\
-shortness: EMS concept\\
-of: none\\
-breath: EMS concept\\
-Lung: EMS concept\\
-transplantation: EMS concept\\
-is: none\\
-an: none\\
-option: none\\

Step2: RefineEMS concept from Step 1 by following criteria,\\
1.concatenate EMS concept spans\\
2.remove extra irrelevant words in EMS concept\\
-cough: EMS concept\\
-wheezing: EMS concept\\
-shortness of breath: EMS concept\\
-Lung transplantation: EMS concept\\

Step3: Return the your result:
["cough", wheezing", "shortness of breath", "Lung transplantation"]

Now is the real text: \textbf{RAW TEXT}
\end{tcolorbox}
  \caption{Prompt for EMS Concept Extraction}
  \label{fig:cpt_extraction}
\end{figure}

\begin{table*}[t!]
  \centering
  \setlength{\tabcolsep}{1mm}  
  \begin{tabular}{@{} l l | l l l | r r | r r | r r @{}}
    \toprule
    \textbf{Data} & \textbf{Comparison} & \textbf{QA} & \textbf{KB} & \textbf{PR} 
      & \textbf{KB $\cap$ QA} & \textbf{PR $\cap$ QA} 
      & \textbf{KB $\setminus$ QA} & \textbf{PR $\setminus$ QA} 
      & \textbf{KB $\cup$ QA} & \textbf{PR $\cup$ QA} \\
    \midrule
    \multirow{3}{*}{Public}
      & Vocab            & 15,892 & 33,965 &  6,773 & 13,183 & 3,359 & 20,782 & 3,414 & 36,674 & 19,306 \\
      & Cpts (w/o norm)  & 25,594 & 61,003 & 9,383 & 10,661 & 2,269 & 50,342 & 7,114 & 75,936 & 32,708 \\
      & Cpts (w norm)    & 17,262 & 34,627 &  8,088 &  10,927 & 2,637 & 23,700 & 5,451 & 40,962 & 22,713 \\
    \midrule
    \multirow{3}{*}{Private}
      & Vocab            & 10,496 & 33,965 &  6,773 &  9,540 & 2,966 & 24,425 & 3,807 & 34,921 & 14,303 \\
      & Cpts (w/o norm)  & 11,621 & 61,003 & 9,383 &  6,180 & 1,669 & 54,823 & 7,714 & 66,444 & 19,335 \\
      & Cpts (w norm)    &  8,689 & 34,627 &  8,088 &  6,299 & 1,969 & 28,328 & 6,119 & 37,017 & 14,808 \\
    \bottomrule
  \end{tabular}
  \caption{Detailed overlap statistics between QA and KB for public and private data.}
  \label{tab:overlap_detailed}
\end{table*}

\begin{table*}[t!]
  \centering
  \setlength{\tabcolsep}{1mm}
  \begin{tabular}{@{}llc@{}}
    \toprule
    \textbf{Resource Type and Title} 
      & \textbf{Level} 
      & \textbf{\# Vocab} \\
    \midrule
    \multicolumn{3}{@{}l}{\textbf{Transcripts}} \\
    \makecell[l]{Emergency Care and Transportation of\\the Sick and Injured Advantage Package~\cite{youtubeCD}}
      & EMT       & 23,836 \\
    \makecell[l]{Nancy Caroline’s Emergency Care in\\the Streets~\cite{youtubeCD}}
      & Paramedic &        \\
    \makecell[l]{AAOS Advanced Emergency Medical\\Technician (AEMT) 4th Ed~\cite{youtubeEMP}}
      & AEMT      &        \\
    AAOS Critical Care Transport Paramedic~\cite{youtubeEMP}
      & Paramedic &        \\
    \midrule
    \multicolumn{3}{@{}l}{\textbf{Textbooks}} \\
    EMT Exam for Dummies~\cite{emtfourdummies}
      & EMT       &  2,025 \\
    \makecell[l]{Emergency Medical Services: Clinical\\Practice and Systems Oversight~\cite{wileyEMS}}
      & —         & 18,799 \\
    \makecell[l]{EMS Essentials: A Resident’s Guide\\to Prehospital Care\cite{emraEssentials}}
      & —         &  4,123 \\
    Emergency Medical Response - RedCross~\cite{redcrossEMR}
      & EMR       & 10,096 \\
    \makecell[l]{Emergency Medical Responder: Your First\\Response in Emergency Care~\cite{firstResponderLA}}
      & EMR       &  7,794 \\
    \midrule
    \multicolumn{3}{@{}l}{\textbf{Guidelines}} \\
    Professional Responder Cheat Sheet~\cite{cheatsheet}
      & EMR       &  1,569 \\
    EMS Pharmacology Reference Guide~\cite{riPharm}
      & —         &  3,072 \\
    Paramedic Medication Manual~\cite{delawarePharm}
      & Paramedic &  2,011 \\
    ODEMSA Regional EMS Documents~\cite{odemsaDocs}
      & —         &  6,340 \\
    \midrule
    \multicolumn{3}{@{}l}{\textbf{Slides}} \\
    \makecell[l]{Emergency Care and Transportation of\\the Sick and Injured Advantage Package~\cite{jbLearning}}
      & EMT       &  8,597 \\
    \midrule
    \multicolumn{3}{@{}l}{\textbf{Flashcards \& Study Guides}} \\
    EMS Study Guides~\cite{appEmtPrep}
      & EMR, EMT, AEMT, Paramedic &  8,373 \\
    \bottomrule
  \end{tabular}
  \caption{Knowledge base collection details}
  \label{tab:kb_emsqa_stats_vocab}
\end{table*}

\subsection{A.3.3\quad UMLS Concept Normalization}
\label{sec:3.3}
We applied UMLS API~\cite{bodenreider2004unified} to normalize the extracted EMS concepts and retained only terms whose semantic types fell into one of the following categories: Sign or Symptom (Sosy), Finding (Fndg), Laboratory or Test Result (Lbtr), Clinical Attribute (Clna), Quantitative Concept (Qnco), Qualitative Concept (Qlco), Disease or Syndrome (Dsyn), Mental or Behavioral Dysfunction (Mobd), Pathologic Function (Patf), Neoplastic Process (Neop), Congenital Abnormality (Cgab), Anatomical Abnormality (Anab), Injury or Poisoning (Inpo), Cell or Molecular Dysfunction (Celf), Body Part/Organ/Organ Component (Bpoc), Body Location or Region (Bodl), Body Space or Junction (Bsoj), Body System (Bodsys), Therapeutic or Preventive Procedure (Topp), Diagnostic Procedure (Diap), Laboratory Procedure (Lbpr), Imaging Procedure (Impr), Health Care Activity (Hlca), Clinical Drug (Clnd), Pharmacologic Substance (Phsu), Antibiotic (Antb), Vitamin (Vita), Organic Chemical (Orch), Amino Acid/Peptide/Protein (Aapp), Biologically Active Substance (Bacs), Hazardous or Poisonous Substance (Hops), Steroid (Strd), Hormone (Horm), Medical Device (Medd), Temporal Concept (Tmco), Spatial Concept (Spco), and Functional Concept (Fngp).

\subsection{A.3.4\quad Knowledge and EMSQA Overlap Statistics}
\label{sec:3.4}
The detailed statistics such as the total number of concepts in EMSQA and KB, overlapped concepts (KB $\cap$ QA), union concepts (KB $\cup$ QA), KB unique concepts (KB $\setminus$ QA) are shown in Table~\ref{tab:overlap_detailed}. The syntactic (hit rate) reported in paper is calculated by KB $\cap$ QA / QA.


\subsection{A.3.5\quad Knowledge Base Collection Details}
\label{sec:3.5}
Table~\ref{tab:kb_emsqa_stats_vocab} shows the detailed information of every knowledge source we collected to construct external EMS knowledge bases. Specifically, we list the resource type, title, URL, certification levels, and vocabulary of each source. In total, there are 2,545,192 tokens in our external knowledge bases, and	the unique vocabulary size is 34,110.

\section{A.4\quad NEMSIS Patient Care Report Preprocessing \& Statistics}
\label{sec:4}
Access to NEMSIS 2021 Public-Release Research Dataset requires submitting a request on the NEMSIS website\footnote{https://nemsis.org/using-ems-data/request-research-data/}.

\subsection{A.4.1\quad NEMSIS Patient Record Preprocessing}
\label{sec:4.1}
To concatenate information for each patient record in the NEMSIS dataset, we used the following fields in different tables and concatenated them by the shared primary key. The keys in our patient records include "Date - Time of Symptom Onset", "Chief Complaints", "Possible Injury", "Cause of Injury", "Chief Complaint Anatomic Location", "Chief Complaint Organ System", "Gender", "Age", "Age Unit", "Level of Responsiveness (AVPU)", "Primary Symptoms", "Other Associated Symptoms", "Primary Impressions", "Secondary Impressions", "Protocol Age Category", "Protocols", "Date - Time Vital Signs Taken", "ECG Type", "SBP (Systolic Blood Pressure)", "Heart Rate", "Respiratory Rate", "Pulse Oximetry", "Blood Glucose Level", "End Tidal Carbon Dioxide (ETCO2)", "Glasgow Coma Score-Eye", "Glasgow Coma Score-Verbal", "Glasgow Coma Score-Motor", "Pain Scale Score", "Stroke Scale Score", "Stroke Scale Type", "Reperfusion Checklist", "Date - Time Medication Administered", "Medication Administered Prior to this Unit's EMS Care", "Medication Given", "Medication Dosage", "Medication Dosage Units", "Response to Medication", "Role - Type of Person Administering Medication", "Date - Time Procedure Performed", "Procedure", "Number of Procedure Attempts", "Response to Procedure", "Role - Type of Person Performing the Procedure", "Alcohol - Drug Use Indicators", "Barriers to Patient Care", "Cardiac Arrest", "Date - Time of Cardiac Arrest", "First Monitored Arrest Rhythm of the Patient", "Cardiac Arrest Etiology", "Type of CPR Provided", "Cardiac Rhythm on Arrival at Destination", "Reason CPR - Resuscitation Discontinued", "Incident - Patient Disposition", "EMS Transport Method", "Transport Mode from Scene", "Initial Patient Acuity", "Final Patient Acuity".

We further classify patient records to seven subject areas by the field protocol EMS. Specifically, 

\textbf{"Assessment"} includes "General-Universal Patient Care/ Initial Patient Contact", "General-Individualized Patient Protocol"

\textbf{Medical \& OB} includes "Medical-Seizure", "Medical-Nausea/Vomiting", "Medical-Influenza-Like Illness/ Upper Respiratory Infection", "Medical-Abdominal Pain", "Medical-Altered Mental Status", "OB/GYN-Pregnancy Related Emergencies", "Medical-Supraventricular Tachycardia (Including Atrial Fibrillation)", "Medical-Cardiac Chest Pain", "Medical-Tachycardia", "Medical-Syncope", "Medical-Stroke/TIA", "Medical-Respiratory Distress/Asthma/COPD/Reactive Airway", "Medical-Respiratory Distress-Bronchiolitis", "Medical-Diarrhea", "Medical-Hypoglycemia/Diabetic Emergency", "Medical-Hyperglycemia", "Medical-Allergic Reaction/Anaphylaxis", "Medical-Hypertension", "Medical-Bradycardia", "Medical-Pulmonary Edema/CHF", "Medical-Hypotension/Shock (Non-Trauma)", "Medical-Opioid Poisoning/Overdose", "Medical-ST-Elevation Myocardial Infarction (STEMI)", "Medical-Ventricular Tachycardia (With Pulse)", "Medical-Stimulant Poisoning/Overdose", "Medical-Respiratory Distress-Croup", "Medical-Adrenal Insufficiency", "Medical-Beta Blocker Poisoning/Overdose", "Medical-Calcium Channel Blocker Poisoning/Overdose", "OB/GYN-Gynecologic Emergencies", "OB/GYN-Childbirth/Labor/Delivery", "OB/GYN-Eclampsia", "OB/GYN-Post-partum Hemorrhage", "General-Overdose/Poisoning/Toxic Ingestion", "General-Fever", "General-Epistaxis", "General-Back Pain", "General-Pain Control", "Environmental-Altitude Sickness", "Environmental-Cold Exposure", "Environmental-Hypothermia", "Environmental-Heat Stroke/Hyperthermia", "Environmental-Heat Exposure/Exhaustion"

\textbf{"Airway"} includes "Airway", "Airway-Failed", "Airway-Sedation Assisted (Non-Paralytic)", "Airway-Rapid Sequence Induction (RSI-Paralytic)", "Airway-Obstruction/Foreign Body",

\textbf{"EMS Operations"} includes "Exposure-Airway/Inhalation Irritants", "General-Neglect or Abuse Suspected", "Exposure-Explosive/ Blast Injury", "General-IV Access",
"General-Refusal of Care", "General-Behavioral/Patient Restraint", "General-Interfacility Transfers", "General-Spinal Immobilization/Clearance", "General-Medical Device Malfunction", "General-Law Enforcement - Assist with Law Enforcement Activity", "General-Extended Care Guidelines", "General-Exception Protocol", "General-Indwelling Medical Devices/Equipment", "General-Community Paramedicine / Mobile Integrated Healthcare", "General-Law Enforcement - Blood for Legal Purposes", "General-Dental Problems", "Exposure-Biological/Infectious", "Exposure-Carbon Monoxide"", "Exposure-Chemicals to Eye", "Exposure-Smoke Inhalation", "Exposure-Cyanide", "Exposure-Radiologic Agents", "Exposure-Blistering Agents", "Exposure-Nerve Agents"

\textbf{"Cardiovascular"} includes "General-Cardiac Arrest", "Cardiac Arrest-Asystole", "Cardiac Arrest-Determination of Death / Withholding Resuscitative Efforts", "Cardiac Arrest-Pulseless Electrical Activity", "Cardiac Arrest-Do Not Resuscitate", "Cardiac Arrest-Ventricular Fibrillation/ Pulseless Ventricular Tachycardia", "Cardiac Arrest-Post Resuscitation Care", "Cardiac Arrest-Special Resuscitation Orders", "Cardiac Arrest-Hypothermia-Therapeutic"

\textbf{"Pediatrics"} includes "Medical-Apparent Life Threatening Event (ALTE)", "Medical-Newborn/ Neonatal Resuscitation"

\textbf{"Trauma"} includes "Environmental-Frostbite/Cold Injury", "Injury-Head", "Injury-Extremity", "Injury-Burns-Thermal", "Injury-General Trauma Management", "Injury-Multisystem", "Injury-Eye", "Injury-Mass/Multiple Casualties", "Injury-Amputation", "Injury-Spinal Cord", "Injury-Conducted Electrical Weapon (e.g., Taser)", "Injury-Bleeding/ Hemorrhage Control", "Injury-Bites and Envenomations-Land", "Injury-Facial Trauma", "Injury-Cardiac Arrest", "Injury-Crush Syndrome", "Injury-Thoracic", "Injury-Drowning/Near Drowning", "Injury-Diving Emergencies", "Injury-Electrical Injuries", "Injury-Topical Chemical Burn", "Injury-Impaled Object", "Injury-Bites and Envenomations-Marine", "Injury-Lightning/Lightning Strike", "Injury-SCUBA Injury/Accidents".

\begin{table}[ht]
\centering
\begin{tabular}{@{}l r@{}}
\toprule
\multicolumn{2}{c}{\textbf{NEMSIS Patient Records}} \\
\midrule
Total cases              & 4,003,430       \\
Total tokens             & 1,247,811,207   \\
Average tokens/case    & 311.7           \\
Min tokens               & 202             \\
Max tokens               & 824             \\
Vocabulary               & 6,838           \\
\bottomrule
\end{tabular}
\caption{Statistics for the NEMSIS patient care reports}
\label{tab:nemsis_patient_records}
\end{table}

\subsection{A.4.2\quad NEMSIS Patient Record Statistics}
\label{sec:4.2}
The detailed statistics of processed NEMSIS data is shown at Table~\ref{tab:nemsis_patient_records}. There are in total 4,003,430 patient records with average 311.7 tokens per case. The total number vocabulary is 6,838.

We also show the statistics such as the total number of concepts in EMSQA and PR, overlapped concepts (PR $\cap$ QA), union concepts (PR $\cup$ QA), PR unique concepts (PR $\setminus$ QA) are shown in Table~\ref{tab:overlap_detailed}. The syntactic (hit rate) reported in paper is calculated by PR $\cap$ QA / QA.

\section{A.5\quad Human Evaluation of EMSQA and Knowledge Base}
\label{sec:5}
To validate the quality and clinical utility of our EMSQA benchmark and its accompanying curated knowledge bases, we conducted an expert review on a randomly sampled subset of 100 questions. Sampling was stratified to ensure balanced coverage across certification levels—25 questions each for EMR, EMT, AEMT, and Paramedic—and across five major medical subject areas—20 questions each for Airway, Trauma, Cardiology, Medical, and Operations—while also drawing uniformly from all source links.

We asked one EMT-certified professional to rate each question–answer pair on four dimensions:
\begin{description}
  \item[\textbf{Clinical Accuracy:}]  
    Verify that the provided “correct” answer is clinically sound.  
    \emph{Rating:} Yes / No

  \item[\textbf{Difficulty Level:}]  
    Judge whether the question’s difficulty matches the expected level for its certification.  
    \emph{Rating:} 1 (Strongly Disagree) to 5 (Strongly Agree)

  \item[\textbf{Subject Area Accuracy:}]  
    Confirm that the assigned medical subject area appropriately reflects the question’s content.  
    \emph{Rating:} 1 (Strongly Disagree) to 5 (Strongly Agree)

  \item[\textbf{Knowledge Relevance:}]  
    Assess whether the assembled knowledge base sufficiently supports answering the question.  
    \emph{Rating:} 1 (Strongly Disagree) to 5 (Strongly Agree)
\end{description}

Table~\ref{tab:expert_eval_results} shows average rating for four dimensions. The results indicate that: (1) provided answers are clinically accurate; (2) the EMT expert largely agrees with the assigned certification levels and subject areas; and (3) the curated knowledge base is helpful for answering the questions.
\begin{table}[t]
  \centering
  \small
  \setlength{\tabcolsep}{1mm}
  \begin{tabular}{@{}l l c@{}}
    \toprule
    \textbf{Dimension}    &\textbf{Scale}    & \textbf{Expert Rating}            \\
    \midrule
    Clinical Accuracy     & 0-100   &                       100\\
    Difficulty Level & 1-5         &  3.95\\
    Subject Area Accuracy & 1-5       &  4.85\\
    Knowledge Relevance & 1-5     &  4.39\\
    \bottomrule
  \end{tabular}
  \caption{Human Expert evaluation on EMSQA \& KB.}
  \label{tab:expert_eval_results}
\end{table}

\section{A.6 \quad Retrieval Strategy Performance}
\label{sec:6}
In the paper we present three different retrieval strategies (\textbf{Global}, \textbf{Filter then Retrieve (FTR)}, \textbf{Retrieve then Filter (RTF)}). To evaluate the retrieval performance, we uniformly sampled 100 questions from the 4 different certification levels (25 questions per certification level) and 10 subject areas ($\sim$ 10 questions per subject area). For each question, we use MedCPT to retrieve 32 documents from Knowledge Bases.

Since we do not have the ground-truth question-document pairs, we use LLM-as-a-judge~\cite{zheng2023judging} to evaluate the \textbf{``Relevance''}~\cite{es2024ragas, asai2023self} and~\textbf{``Supportiveness''}~\cite{asai2023self} of the 32 retrieved documents for every question. Relevance judges if each retrieved chunk is pertinent to the question, while supportiveness judges if this retrieved chunk helps answer the question.

\begin{figure}[b!]
  \centering
    \begin{tcolorbox}[breakable=false,width=\columnwidth,]
    \label{zero-shot}
\textbf{PROMPT:}\\
Question: \{\}\\
Answer: \{\}\\
Retrieved Chunk: \{\}\\
Rubric\\
- Relevance\\
• "Relevant" — The chunk provides information that is useful for answering the question.  \\
• "Irrelevant" — The chunk is off-topic or supplies no helpful information.\\
- Supportiveness\\
Evaluate how much of a hypothetical answer could be *entailed* by this chunk alone.  \\
• "Fully" — All necessary information is present; the answer would be completely supported.  \\
• "Partially" — Some but not all required information is present.  \\
• "None" — No information is supported, or the chunk contradicts the answer.
\end{tcolorbox}
\caption{Prompt for LLM-as-a-judge}
\label{fig:llm-as-a-judge}
\end{figure}

As shown in Figure~\ref{fig:llm-as-a-judge}, we present the prompt template we used in LLM-as-a-judge. To evaluate “Relevance” and “Supportiveness” we use four Information Retrieval metrics: Top-$k$ Hit rate ($\mathit{hit}@k$), which measures the fraction of questions for which at least one positive chunk appears in the top-$k$ retrieved; Top-$k$ Precision ($p@k$), which computes the average proportion of positive chunks among the top-$k$ for each question; Mean Reciprocal Rank ($\mathrm{MRR}$), which takes the reciprocal of the rank position of the first positive chunk for each question and then averages these reciprocals—thereby reflecting how early in the ranking the first positive appears; and Mean Average Precision ($\mathrm{MAP}$), which for each question computes the average of precisions at every positive chunk’s position and then averages those per‐question APs, capturing both the ranking quality and distribution of positives over the entire list. We omit Top-$k$ Recall ($r@k$) since we lack ground-truth documents.

For each question $q$ in sampled questions $\mathcal{Q}$ and retrieved chunk $i$, we define:
\begin{equation}
r_{q,i}^{\mathrm{rel}} =
\begin{cases}
1 & \text{if chunk $i$ is Relevant},\\
0 & \text{otherwise},
\end{cases}
\end{equation}

\begin{equation}
r_{q,i}^{\mathrm{sup}} =
\begin{cases}
1 & \text{if chunk $i$ is Fully or Partially support},\\
0 & \text{otherwise}.
\end{cases}
\end{equation}

Then, for either $\{r_{q,i}^{\mathrm{rel}}\}$ or $\{r_{q,i}^{\mathrm{sup}}\}$:

\begin{equation}
\mathit{hit}@k = \frac{1}{|\mathcal{Q}|}
\sum_{q\in\mathcal{Q}}
\mathbf{1}\!\Bigl(\sum_{i=1}^k r_{q,i} > 0\Bigr)
\end{equation}

\begin{equation}
p@k = \frac{1}{|\mathcal{Q}|}
\sum_{q\in\mathcal{Q}}
\frac{1}{k}\sum_{i=1}^k r_{q,i}
\end{equation}

\begin{equation}
\mathrm{MRR} = \frac{1}{|\mathcal{Q}|}
\sum_{q\in\mathcal{Q}}
\frac{1}{\min\{\,i : r_{q,i}=1\}}
\end{equation}

\begin{equation}
\mathrm{MAP} = \frac{1}{|\mathcal{Q}|}
\sum_{q\in\mathcal{Q}}
\frac{1}{\sum_{i=1}^N r_{q,i}}
\sum_{i=1}^N r_{q,i}\,\frac{1}{i}\sum_{j=1}^i r_{q,j}\,.
\end{equation}


\begin{table}[t!]
  \centering
  \small
  \setlength{\tabcolsep}{1mm}
  \begin{tabular}{@{}lcccc@{}}
    \toprule
    \multicolumn{5}{c}{\textbf{Relevance}} \\
    \textbf{Strategy} 
      & \textbf{Hit@k} & \textbf{P@k} & \textbf{MRR} & \textbf{MAP} \\
    & (1 / 5 / 10) & (1 / 5 / 10) & & \\
    \midrule
    Global               & 44.0 / 69.0 / 78.0 & 44.0 / 35.0 / 31.7 & 55.6 & 39.4 \\
    FTR & 44.0 / 71.0 / 80.0 & 44.0 / 35.0 / 32.2 & \textbf{57.6} & 39.6 \\
    RTF & \textbf{45.0} / \textbf{75.0} / \textbf{81.0} & \textbf{45.0} / \textbf{35.0} / \textbf{32.5} & 57.5 & \textbf{39.6} \\
    \midrule
    \multicolumn{5}{c}{\textbf{Supportiveness}} \\
    \textbf{Strategy} 
      & \textbf{Hit@k} & \textbf{P@k} & \textbf{MRR} & \textbf{MAP} \\
    & (1 / 5 / 10) & (1 / 5 / 10) & & \\
    \midrule
    Global               & 25.0 / 52.0 / 67.0 & 25.0 / 21.0 / 20.3 & 38.3 & 27.6 \\
    FTR & 27.0 / \textbf{59.0} / \textbf{70.0} & 27.0 / \textbf{24.0} / 21.2 & 39.7 & \textbf{29.2} \\
    RTF & \textbf{29.0} / 58.0 / 70.0 & \textbf{29.0} / 23.0 / \textbf{21.6} & \textbf{39.8} & 28.8 \\
    \bottomrule
  \end{tabular}
  \caption{Retrieval Relevance and Supportiveness}
  \label{tab:retrieval_results}
\end{table}
Table~\ref{tab:retrieval_results} presents the performance of three retrieval strategies,Global (direct retrieval), Retrieve-then-Filter (RTF) and Filter-then-Retrieve (FTR), in terms of both “Relevance” and “Supportiveness”. In terms of ``Relevance'', RTF and FTR show slightly better performance than Global across all metrics, indicating that the proposed strategies retrieve relevant chunks earlier and more reliably. In terms of ``Supportiveness'', both RTF and FTR achieve higher Hit@k and P@k and slightly better MRR and MAP than Global, suggesting they more effectively capture fully or partially supportive evidence. Overall, although based on a small study, these results show that the proposed retrieval strategies improve the extraction of both relevant and supportive context from the knowledge base, providing stronger grounding for the LLM’s final answer. A more comprehensive evaluation of retrieval strategies is beyond the scope of this paper and left for future work.


\section{A.7\quad Benchmark LLMs}
\label{sec:7}
\subsection{A.7.1\quad Zero-shot Performance per Subject Area}
\label{sec:7.1}
\begin{figure}[hbp]
  \centering
  \includegraphics[width=0.7\columnwidth]{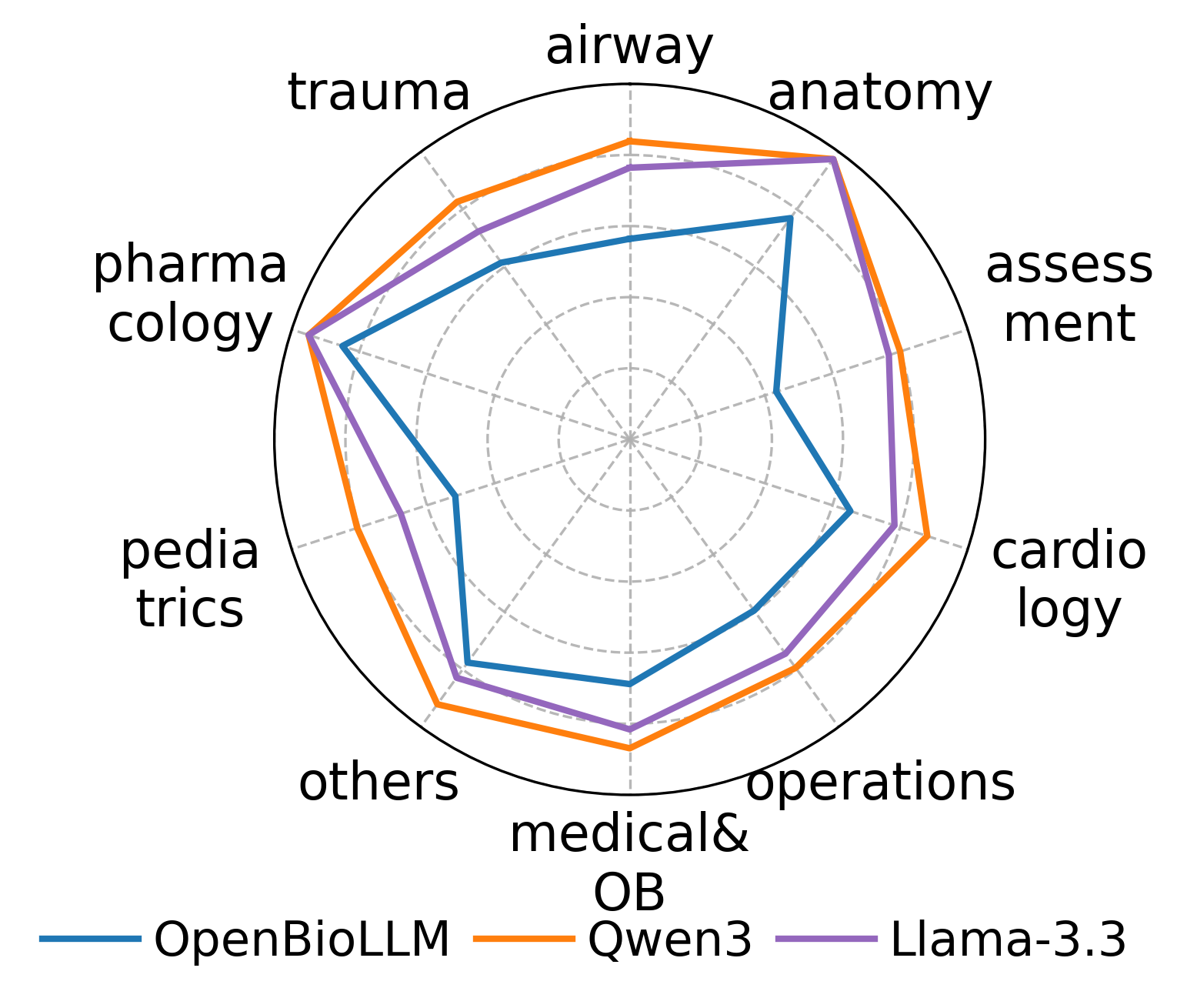}
  \caption{Zero-shot Performance per Subject Area on EMSQA-Private Dataset}
  \label{fig:private-radar}
\end{figure}

As shown in Figure~\ref{fig:private-radar}, we show the zero-shot LLMs' per subject area performance on private dataset. We only run open-source models on our private dataset. The conclusion remains consistent on the private dataset: \textbf{LLMs excel in pharmacology and anatomy but falter on core NREMT domains}. Models answer pharmacology and anatomy items reliably, yet stumble on ``trauma'', ``airway management'', ``EMS operations'', and ``medical\&OB'' questions. One reason may be task complexity: the former are largely single-hop, fact-based queries, whereas the latter demand multi-hop reasoning and richer EMS context.

\subsection{A.7.2\quad Detailed statistics of Benchmarks}
\label{sec:7.2}
Table \ref{tab:emsqa_zeroshot_benchmark} summarizes the full benchmarking results for each model and prompting strategy, reporting accuracy and F1 across all four EMS certification levels (EMR, EMT, AEMT, Paramedic) as well as overall performance. Below are several key findings:

\textbf{Closed-source models outperform open-source models.} In particular, OpenAI-o3 consistently achieves the highest overall accuracy of 92.39. Among open-source models, Qwen3-32B achieves the best accuracy of 85.70, though a significant gap remains compared to closed-source models.

\textbf{Few-shot prompting improves accuracy up to a point.} We varied the number of in‑context exemplars from 0 to 64 and observed that incorporating a small number of examples yields substantial gain over the zero‑shot baseline. However, adding examples beyond a certain point leads to diminishing or no improvement.

\textbf{Expert-CoT helps guide LLM reasoning.} Integrating domain expert knowledge via CoT‑Expert guides reasoning towards appropriate context and consistently boosts CoT prompting performance by up to 2.05\% across models. Also, using the Filter's predicted expertise for Expert-CoT leads to comparable performance to when using ground-truth.

\begin{table*}[t!]
  \centering
  \small
  \setlength{\tabcolsep}{1mm}
  \begin{tabular*}{0.95\textwidth}{@{\extracolsep{\fill}} l c *{5}{c c}}
    \toprule
    & & \multicolumn{10}{c}{\textbf{EMSQA—Public}} \\
    \cmidrule(lr){3-12}
    \textbf{Model} & \textbf{Prompt}
      & \multicolumn{2}{c}{\textbf{EMR}}
      & \multicolumn{2}{c}{\textbf{EMT}}
      & \multicolumn{2}{c}{\textbf{AEMT}}
      & \multicolumn{2}{c}{\textbf{Paramedic}}
      & \multicolumn{2}{c}{\textbf{Overall}} \\
    \cmidrule(lr){3-4} \cmidrule(lr){5-6}
    \cmidrule(lr){7-8} \cmidrule(lr){9-10} \cmidrule(lr){11-12}
    & 
      & \textbf{Acc.} & \textbf{F1}
      & \textbf{Acc.} & \textbf{F1}
      & \textbf{Acc.} & \textbf{F1}
      & \textbf{Acc.} & \textbf{F1}
      & \textbf{Acc.} & \textbf{F1} \\
    \midrule
    \multirow{8}{*}{Qwen3-32B}
     & 0-shot   & 79.08 & 79.08 & 82.98 & 82.98 & 83.68 & 83.68 & 86.69 & 86.69 & 83.55 & 83.55 \\
     & 4-shot   & 80.89 & 80.89 & 84.31 & 84.31 & 86.05 & 86.05 & 86.06 & 86.06 & 84.41 & 84.41 \\
     & 8-shot   & 80.20 & 80.20 & 84.60 & 84.60 & 85.79 & 85.79 & 86.14 & 86.14 & 84.39 & 84.39 \\
     & 16-shot  & 81.03 & 81.03 & 84.23 & 84.23 & 85.79 & 85.79 & \textbf{87.31} & 87.31 & 84.82 & 84.82 \\
     & 32-shot  & 77.68 & 77.68 & 81.13 & 81.13 & 81.58 & 81.58 & 83.01 & 83.01 & 81.13 & 81.13 \\
     & 64-shot  & 80.33 & 80.33 & 80.09 & 80.09 & 85.26 & 85.26 & 85.43 & 85.43 & 82.48 & 82.48 \\
     & CoT      & 80.06 & 80.06 & 82.90 & 82.95 & 85.26 & 85.26 & 86.92 & 86.92 & 84.96 & 84.97 \\
     \multirow{1}{*}{\quad(GT)}& Expert-CoT    & \textbf{82.43} & \textbf{82.48}  & \textbf{85.49} & \textbf{85.49}  & \textbf{87.89} & \textbf{87.89}  & 87.16 & 87.16  & \textbf{85.70} & \textbf{85.71} \\
     \multirow{1}{*}{\quad(Filter)} & Expert-CoT & 81.52 & 81.51 & 82.24 & 83.68 & 86.37 & 86.65 & 86.98 & \textbf{87.51} & 85.07 & 85.10\\
    \midrule
    \multirow{3}{*}{Llama-3.3}
      & 0-shot   & 79.08 & 79.08 & \textbf{81.05} & 81.05 & \textbf{85.53} & \textbf{85.53} & 85.67 & 85.67 & 81.69 & 82.69 \\
     & CoT  & 79.64 & \textbf{81.29}  & 79.57 & 80.79  & 83.42 & 84.51  & 85.20 & 86.13  & 81.89 & 83.08 \\
     \multirow{1}{*}{\quad(GT)}& Expert-CoT  & 79.22 & 80.32  & 80.68 & \textbf{81.77}  & 83.68 & 85.15  & 85.75 & 86.24  & \textbf{82.42} & \textbf{83.35} \\
     \multirow{1}{*}{\quad(Filter)} & Expert-CoT & \textbf{80.20} & 81.24 & 80.24 & 81.16 & 82.63 & 83.51 & \textbf{85.83} & \textbf{86.27} & 82.40 & 83.18\\
    \midrule
    \multirow{3}{*}{OpenBioLLM}
     & 0-shot   & 51.60 & 51.73 & 54.18 & 54.26 & 61.58 & 61.71 & 63.66 & 63.74 & 57.67 & 57.76 \\
     & CoT   & 53.02 & 53.03 & 54.78 & 54.96 & 62.34 & 62.56 & 63.81 & 63.97 & 59.88 & 60.34 \\
     \multirow{1}{*}{\quad(GT)} & Expert-CoT   & \textbf{55.31} & \textbf{55.89} & \textbf{57.82} & \textbf{57.87} & 64.66 & 65.03 & \textbf{65.77} & \textbf{65.81} & \textbf{61.92} & \textbf{62.03} \\
     \multirow{1}{*}{\quad(Filter)} & Expert-CoT   & 54.91 & 54.91 & 55.12 & 55.12 & \textbf{65.16} & \textbf{65.17} & 65.52 & 65.54 & 61.32 & 61.93 \\
     \midrule
    OpenAI-o3  & 0-shot   & \textbf{90.79} & \textbf{90.79} & \textbf{93.49} & \textbf{93.49} & \textbf{92.37} & \textbf{92.37} & \textbf{92.17} & \textbf{92.17} & \textbf{92.39} & \textbf{92.39} \\
    Gemini-2.5 & 0-shot   & 87.59 & 87.59 & 89.93 & 89.93 & 90.26 & 90.26 & 89.51 & 89.51 & 89.36 & 89.36 \\
    \midrule
    & & \multicolumn{10}{c}{\textbf{EMSQA—Private}} \\
    \cmidrule(lr){3-12}
    \multirow{8}{*}{Qwen3-32B}
    & 0-shot   & 84.81 & 84.93 & 84.55 & 85.28 & 85.55 & 86.47 & 85.27 & 86.09 & 85.11 & 85.89 \\
    & 4-shot   & 86.60 & 86.72 & 85.41 & 85.95 & 86.60 & 87.30 & 85.54 & 86.24 & 85.48 & 86.13 \\
     & 8-shot   & 85.26 & 85.39 & 84.62 & 85.20 & 85.91 & 86.64 & 85.71 & 86.42 & 85.39 & 86.07 \\
     & 16-shot  & 85.78 & 85.95 & 84.98 & 85.55 & 86.11 & 86.84 & 85.68 & 86.41 & 85.52 & 86.19 \\
    & 32-shot  & 82.58 & 82.95 & 81.80 & 82.84 & 83.69 & 85.12 & 82.92 & 84.26 & 82.22 & 83.41 \\
    & 64-shot  & 87.71 & 87.90 & 85.75 & 86.74 & 87.52 & 88.89 & 87.13 & 88.19 & 86.22 & 87.26 \\
    & CoT      & 91.81 & 92.01 & 87.86 & 89.08 & 90.11 & 91.66 & 89.83 & 91.24 & 88.78 & 90.13 \\
     \multirow{1}{*}{\quad(GT)}& Expert-CoT  & \textbf{94.42} & \textbf{94.74}  & \textbf{88.63} & 89.85  & \textbf{92.01} & \textbf{93.58}  & \textbf{91.59} & \textbf{92.83}  & \textbf{89.73} & 90.98 \\
     \multirow{1}{*}{\quad(Filter)} & Expert-CoT & 91.66 & 92.97 & 88.42 & \textbf{90.05} & 91.60 & 93.08 & 91.11 & 92.60 & 89.50 & \textbf{91.20}\\
    \midrule
    \multirow{3}{*}{Llama-3.3}
    & 0-shot     & 75.95 & 76.12 & 77.93 & 78.63 & 76.87 & 77.66 & 76.75 & 77.47 & 78.06 & 78.77 \\
    & CoT        & 87.57 & 88.16 & 84.65 & 85.86 & 86.88 & 88.04 & 85.93 & 87.00 & 85.16 & 86.35 \\
    \multirow{1}{*}{\quad(GT)}& Expert-CoT   & 89.72 & 90.48 & 85.90 & \textbf{87.04} & \textbf{88.66} & \textbf{89.61} & 87.55 & 88.58 & 86.49 & 87.62 \\
    \multirow{1}{*}{\quad(Filter)} & Expert-CoT & \textbf{90.17} & \textbf{90.70} & \textbf{86.06} & 86.98 & 88.41 & 89.47 & \textbf{87.64} & \textbf{88.63} & \textbf{86.63} & \textbf{87.65}\\
    \midrule
    \multirow{3}{*}{OpenBioLLM}
      & 0-shot  & 60.98 & 61.24  & 60.96 & 61.74  & 66.25 & 67.23  & 67.32 & 68.31  & 63.86 & 64.76 \\
     & CoT   & 62.16 & 62.73 & 64.35 & 64.89 & 69.85 & 70.71 & 72.13 & 72.42 & 67.01 & 67.77 \\
     \multirow{1}{*}{\quad(GT)}& Expert-CoT   & \textbf{63.60} & \textbf{63.83} & \textbf{65.01} & \textbf{65.88} & \textbf{70.87} & \textbf{71.84} & \textbf{73.69} & \textbf{74.74} & \textbf{68.75} & \textbf{69.82} \\
     \multirow{1}{*}{\quad(Filter)} & Expert-CoT   & 62.26 & 62.43 & 64.31 & 64.78 & 70.01 & 70.78 & 72.01 & 72.23  & 67.79 & 68.32 \\
    \bottomrule
  \end{tabular*}
  \caption{Zero-/few-shot, CoT and Expert-CoT results for LLMs on EMSQA.}
  \label{tab:emsqa_zeroshot_benchmark}
\end{table*}

\begin{table}[t!]
  \centering
  \small
  \setlength{\tabcolsep}{1mm}
  \begin{tabular}{l l cc cc}
    \toprule
    \textbf{Model} & \textbf{Description} & \multicolumn{2}{c}{\textbf{Public}} & \multicolumn{2}{c}{\textbf{Private}} \\
                   &                 & Acc & F1                           & Acc & F1 \\
    \midrule
    \multirow{4}{*}{Qwen3-32B}     
        & CoT & 84.96 & 84.97 & 88.78 & 90.13\\
        & + Subject Area & 85.36 & 85.48 & 88.94 & 90.61 \\
        & + Certification & 85.40 & 85.45 & 88.89 & 90.62 \\
        & Expert-CoT & 85.70 & 85.71 & 89.73 & 90.98 \\
    \midrule
    \multirow{4}{*}{Qwen3-4B}
        & CoT & 72.35 & 73.09 & 70.58 & 72.02\\
        & + Subject Area & 76.32 & 77.04 & 73.82 & 74.53\\
        & + Certification & 74.53 & 75.37 & 73.29 & 74.31\\
        & Expert-CoT & 77.26 & 78.38 &  74.32 & 75.77\\
    \bottomrule
  \end{tabular}
  \caption{Ablation Study: Effect of Expertise (Certification, Subject Area) on Expert-CoT}
  \label{ablation_study_expertise}
\end{table}

\section{A.8\quad Ablation Study on Certification Level and Subject Area}
\label{sec:8}
To determine which expertise attribute contributes the most to performance, we ablate the two key attributes, ``Subject Area'' and ``Certification'', and measure their individual effects on Expert-CoT. We cannot do the ablation study on Expert-RAG since it only uses the subject area for retrieval. The results are presented in Table~\ref{ablation_study_expertise}. \\
\textbf{Expertise attributes are most effective when used together.} Individually, ``Subject Area’' and ``Certification’' improve performance in Expert-CoT between 0.74-4.91\% in Acc, but combining both attributes yields the best improvements of up to 5.29\% in the 4B model. Besides, ``Subject Area’' helps slightly more than ``Certification’’ in performance improvement. \\
\textbf{Small language models benefit more from expertise guidance}. 
Injecting expertise labels yields only marginal improvements for the 32B LLM model, but leads to significant gains for the 4B model. This disparity suggests that larger models have likely internalized much of the relevant expertise during pre-training, whereas smaller models rely more heavily on explicit expertise guidance to steer their limited capacity toward the correct reasoning trajectory.



\section{A.9 \quad Error Analysis}
\label{sec:9}
\subsection{A.9.1\quad Error Types}
\label{sec:9.1}
\begin{figure}[t!]
  \centering  \includegraphics[width=\columnwidth]{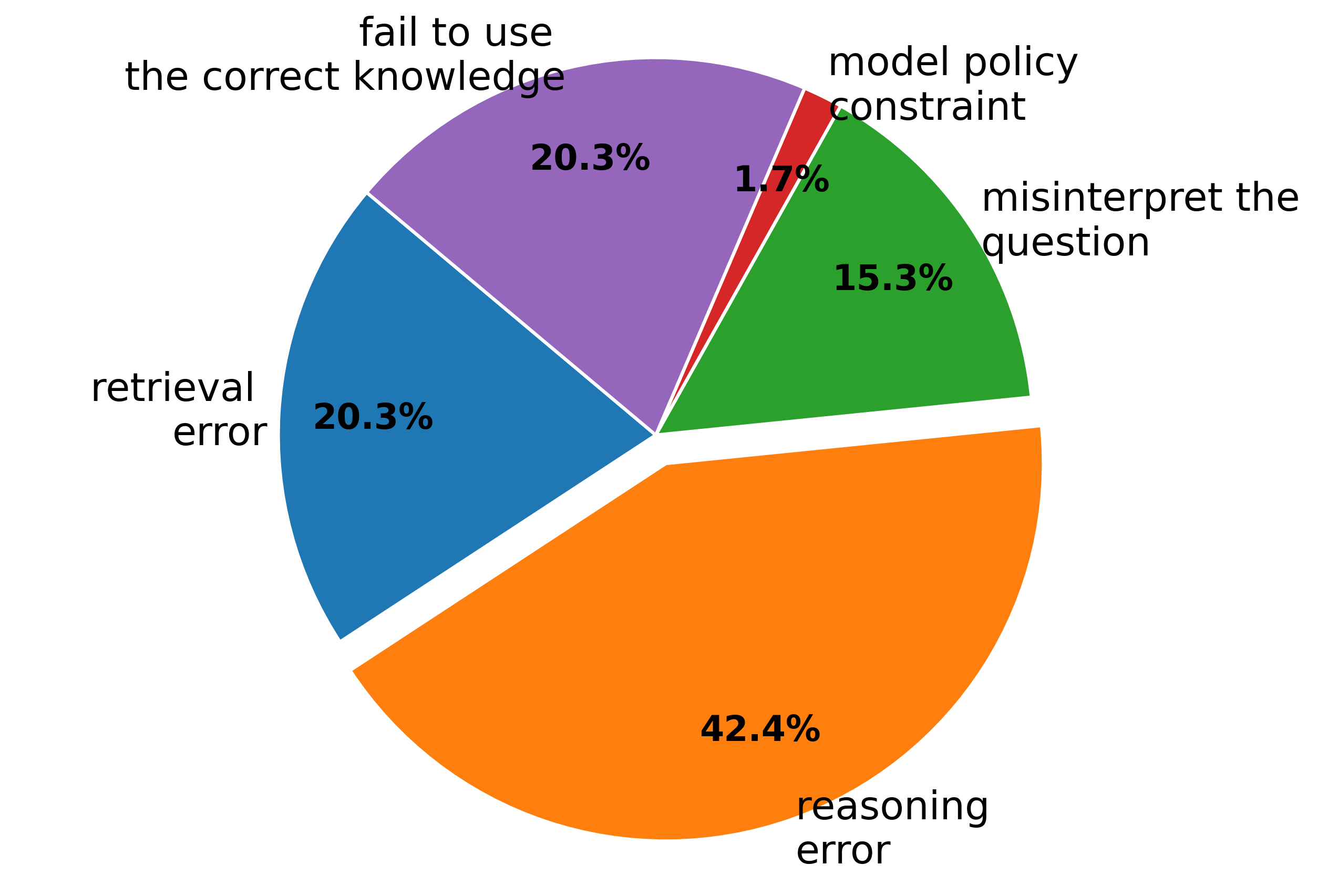}
  \caption{Error Analysis}
  \label{fig:error_analysis}
\end{figure}

We conducted an error analysis on our best 4B model (ExpertRAG-GT + Expert-CoT), which achieved 82.24\% accuracy on EMSQA public split. From the 661 errors on the test set, we randomly selected 50 samples and manually examined each question, its retrieved knowledge, and the model's reasoning process. As shown in Figure~\ref{fig:error_analysis}, we attribute the errors to five main categories:\\
\textbf{1) Reasoning Error} (42.4\%): Retrieved documents were relevant and were used to answer the question, but the model's reasoning was incorrect.\\
\textbf{2) Retrieval Error}(20.3\%): No relevant knowledge was retrieved, causing the model to rely on internal ``model knowledge", which led to incorrect answer.\\
\textbf{3) Knowledge Use Failure} (20.3\%): Part of the retrieved knowledge was relevant to answer the question but the model used other irrelevant knowledge in its reasoning.\\
\textbf{4) Question Misinterpretation} (15.3\%): The model misunderstood the question. For example, for the question \textit{what is the age cutoff for using an AED on a patient?} with choices: \textit{a. 18 year old; b. 5 year old; c. 25 year old; d. 1 year old}, the question refers to  minimum \textit{patient} age (\textit{Correct answer: d. 1 year old}) for routine AED use. However the model interpreted it as the minimum age of the \textit{responder} allowed to operate an AED and incorrectly selected (\textit{a. 18 year old}).\\
\textbf{5) Model Policy Constraint} (1.7\%): The model determined that none of the answer choices were valid and abstained from responding.

\begin{figure}[t!]
  \centering  \includegraphics[width=\columnwidth]{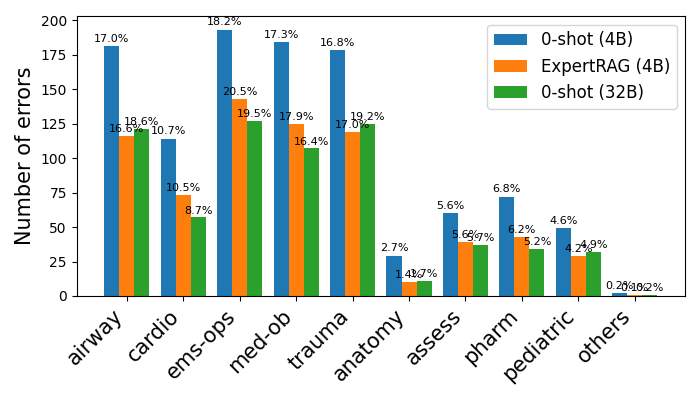}
  \caption{Error Rate per Subject Area}
  \label{fig:error_per_subject_area}
\end{figure}

\begin{figure}[t!]
  \centering  \includegraphics[width=\columnwidth]{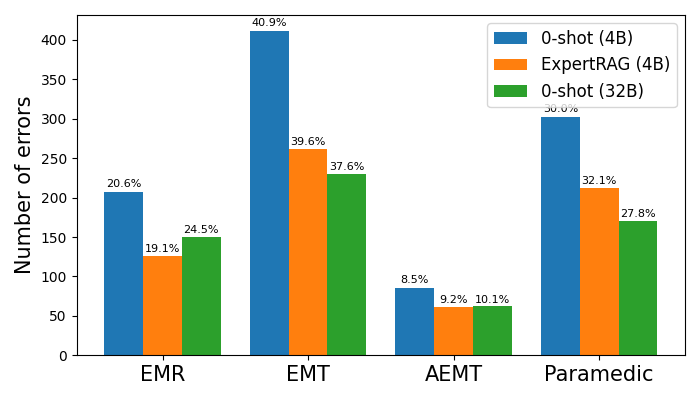}
  \caption{Error Rate per Certification}
  \label{fig:error_per_certification}
\end{figure}

\subsection{A.9.2 \quad Errors per Subject Area and Certification}
\label{sec:9.2}
To investigate how errors are distributed across subject areas and certification levels, we compare the error rates of three models, (Qwen3-4B (0-shot), ExpertRAG-4B (Expert-CoT) and Qwen3-32B (0-shot)). The results are shown in Figure~\ref{fig:error_per_subject_area} and Figure~\ref{fig:error_per_certification}. In subject areas, all three models make most of their mistakes in the core NREMT domains: ``airway'', ``EMS operations'', ``medical–OB/GYN'', ``trauma'', and ``pediatrics''. Together, these categories account for the majority of errors, while ``anatomy'', ``assessment'', ``pharmacology'', and ``others'' contribute relatively few. Across certification levels, all models exhibit the highest error rates at the EMT level, followed by Paramedic and EMR, with AEMT contributing the fewest errors. 

ExpertRAG-4B consistently reduces the absolute number of errors compared to the Qwen3-4B baseline across nearly all subject areas and certification levels. The Qwen3-32B has the fewest errors across three models. One finding is although the total number of errors are reduced in ExpertRAG-4B and Qwen3-32B models, the relative distribution of errors across subject areas is still the same for each model.

\end{document}